\documentclass[runningheads]{llncs}
\usepackage{graphicx}

\usepackage{tikz}
\usepackage{comment}
\usepackage{amsmath,amssymb} 
\usepackage{color}
\usepackage{array}
\usepackage{float}
\usepackage{tabularx}
\usepackage{multicol}
\usepackage{booktabs}
\usepackage{colortbl}
\usepackage{amssymb}
\usepackage{pifont}
\usepackage{tikz}

\usepackage[pagebackref=true,breaklinks=true,colorlinks,bookmarks=false]{hyperref}

\usepackage[noend,ruled,vlined]{algorithm2e}
\usepackage{verbatim}
\usepackage{etoolbox}
\AtBeginEnvironment{algorithm}{\SetArgSty{textrm}}
\SetKwFor{For}{for (}{) $\lbrace$}{$\rbrace$}
\makeatletter
\patchcmd{\@algocf@start}
  {-1.5em}
  {0pt}
  {}{}
\makeatother

\SetCommentSty{mycommfont}

\usepackage{dsfont}

\newif\ifshowedits

\newcommand{\addeditor}[3]{%
  \definecolor{#1color}{rgb}{#3}
  \expandafter\newcommand\csname #1\endcsname[1]{
  \ifshowedits
    {\color{#1color} ##1}
  \else
    {##1}
  \fi
  }%
  \expandafter\newcommand\csname #1rmk\endcsname[1]{
  \ifshowedits
    {\color{#1color} {\bf [#2: ##1]}}
  \else
    {}
  \fi
  }%
}

\newcommand{\newtvar}[1]{
  \expandafter\newcommand\csname #1\endcsname{\text{#1}}
}

\newcommand{\moretextwithfigures}{
\renewcommand{\topfraction}{1}
\renewcommand{\dbltopfraction}{1}
\renewcommand{\bottomfraction}{1}
\renewcommand{\textfraction}{.0}
\renewcommand{\floatpagefraction}{1}
\renewcommand{\dblfloatpagefraction}{1}
}

\newcommand{\calB}{{\cal B}}

\newcommand{\calP}{{\cal P}}

\newcommand{\calS}{{\cal S}}

\newcommand{\calX}{{\cal X}}

\newcommand{\bu}{{\bf u}}
\newcommand{\bv}{{\bf v}}
\newcommand{\bw}{{\bf w}}

\newcommand{\bN}{{\bf N}}

\addeditor{sinisa}{SS}{1., 0.5, 0.0}
\addeditor{vincent}{VL}{0.0, 0.5, 0.0}
\addeditor{michael}{MR}{0.0, 0.0, 0.8}

\moretextwithfigures

\begin{document}


\pagestyle{headings}
\mainmatter
\def\ECCVSubNumber{7803}  

\title{MonteBoxFinder: Detecting and Filtering Primitives to Fit a Noisy Point Cloud}

\titlerunning{MonteBoxFinder}
%
\author{Micha\"el Ramamonjisoa\inst{1} \and
Sinisa Stekovic\inst{2} \and
Vincent Lepetit\inst{1,2}}
\authorrunning{M. Ramamonjisoa et al.}
%
\institute{LIGM, Ecole des Ponts, Univ Gustave Eiffel, CNRS, Marne-la-vall\'ee , France \email{first.lastname@enpc.fr}\\ \and
Institute for Computer Graphics and Vision, Graz University of Technology, Graz,
Austria
\email{sinisa.stekovic@icg.tugraz.at}\\
Project page: \url{https://michaelramamonjisoa.github.io/projects/MonteBoxFinder}}


\maketitle

\begin{figure}[H]
\centering
\setlength{\tabcolsep}{0.5pt}
    \begin{tabular}{cc}
    \includegraphics[trim=550 100 300 320, clip,width=0.48\textwidth]{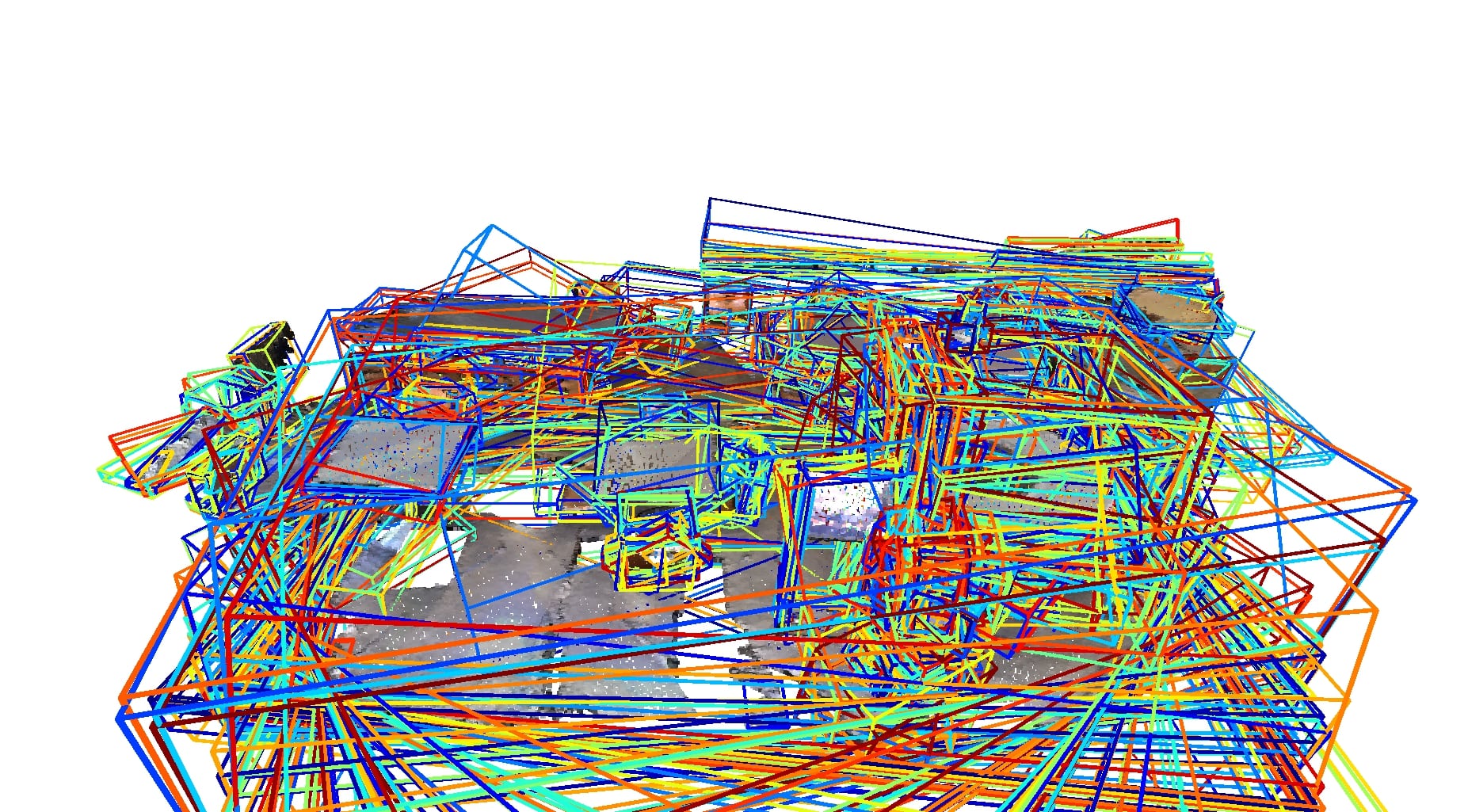} &
    \includegraphics[trim=550 100 300 320, clip,width=0.48\textwidth]{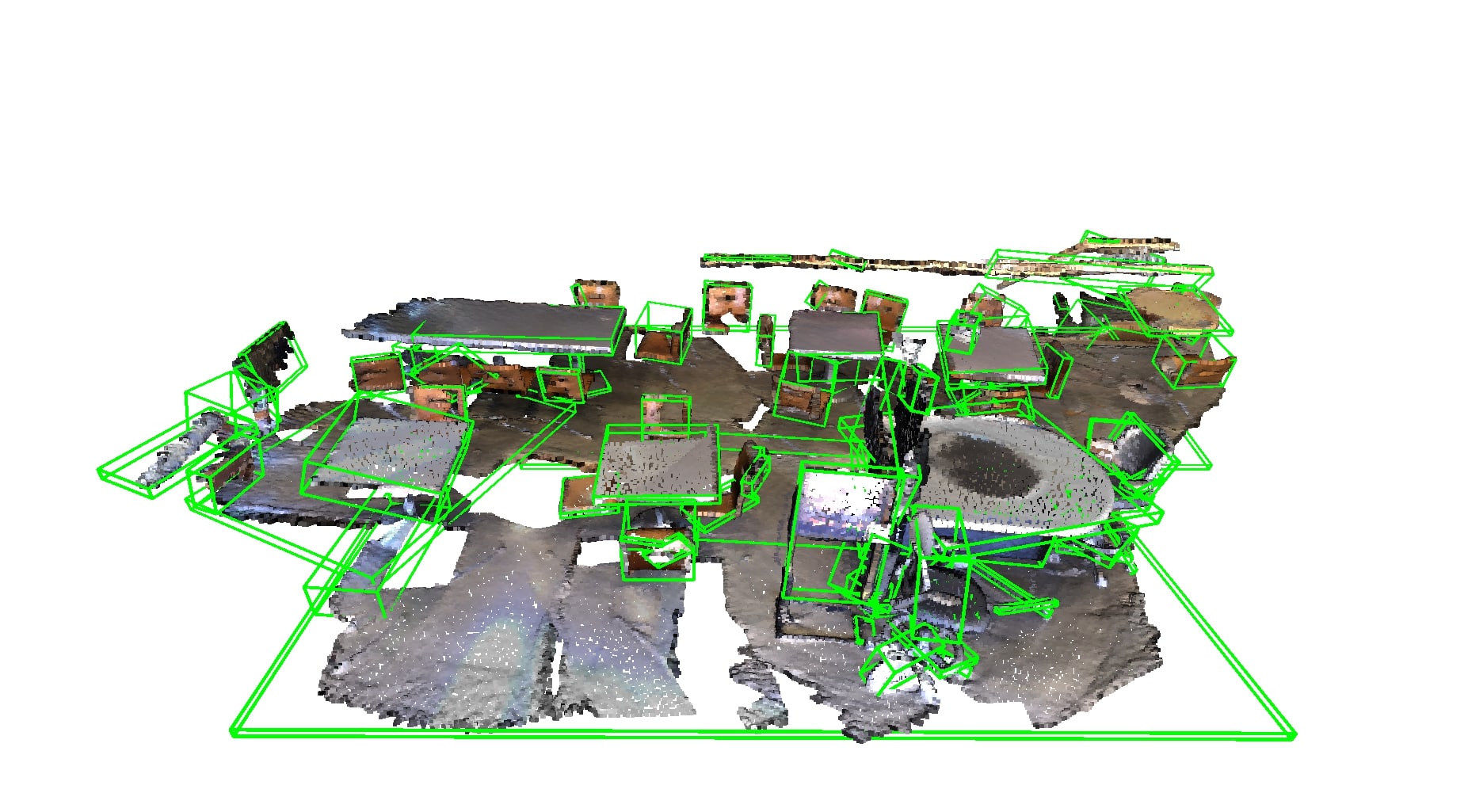} \\
    \end{tabular}\\
    \caption{Given a noisy 3D scan with missing data, our method extracts many possible cuboids, and then efficiently selects the subset that fits the scan best.}
    \label{fig:teaser}
\end{figure}

\begin{abstract}
We present MonteBoxFinder, a method that, given a noisy input point cloud, fits cuboids to the input scene. Our primary contribution is a discrete optimization algorithm that, from a dense set of initially detected cuboids, is able to efficiently filter good boxes from the noisy ones. Inspired by recent applications of MCTS to scene understanding problems, we develop a stochastic algorithm that is, by design, more efficient for our task. Indeed, the quality of a fit for a cuboid arrangement is invariant to the order in which the cuboids are added into the scene. We develop several search baselines for our problem and demonstrate, on the ScanNet dataset, that our approach is more efficient and precise. Finally, we strongly believe that our core algorithm is very general and that it could be extended to many other problems in 3D scene understanding. 
\end{abstract}

\newcommand{\etal}{~\textit{et al}}
\newcommand{\CD}{\textrm{ChamferDistance}}
\newcommand{\ND}{\textrm{CosineDissimilarity}}
\newcommand{\accuracy}{Pr$_\tau$}
\newcommand{\updatesymb}{\textcolor{red}{$\boldsymbol{\rightarrow}$}}
\newcommand{\simulatesymb}{$\boldsymbol{\rightsquigarrow}$}
\newcommand{\terminatesymb}{\textcolor{red}{$\boldsymbol{\times}$}}
\newcommand{\tikzcircle}[2][red,fill=red]{\tikz[baseline=-0.5ex]\draw[#1,radius=#2] (0,0) circle ;}%
\newcommand{\evalsymb}{\tikzcircle[fill=red]{3pt}}

\newcommand{\Cuboid}{\FuncSty{Cuboid} }
\newcommand{\Simulate}{\FuncSty{Simulate} }
\newcommand{\Update}{\FuncSty{Update} }
\newcommand{\loss}{\ell}

\newcommand{\cmark}{\ding{51}}%
\newcommand{\xmark}{\ding{55}}%

\section{Introduction}
Representing a 3D scene with a set of simple geometric primitives is a long-standing computer vision problem~\cite{roberts1963BlocksWorld}. Solving it would provide a light representation of 3D scenes that is arguably easier to  exploit by many downstream applications than a 3D point cloud for example. But maybe more importantly, this would also demonstrate the ability to reach a ``high-level understanding'' of the  scene's geometry, by creating a drastically simplified representation.

In this work, we start from a point cloud of a indoor scene, which can be obtained by 3D reconstruction from images or scanning with an RGB-D camera. Recent works have considered representing 3D point clouds with primitives~\cite{groueix2018,deprelle2019learning,Superquadrics2019CVPR,Paschalidou2020CVPR}; however they consider ``ideal 3D input data'', in the sense that the point cloud is complete and noise-free. By contrast, point clouds from 3D reconstruction or scans are typically very noisy with missing data, and robust methods are required to handle this real data.

To be robust to noise and missing data, we propose a discrete optimization-based method. Our approach does not require any training data, which would be very cumbersome to create manually. Given a point cloud, we extract a large number of primitives. While in our experiments we consider only cuboids as our primitives, our approach can be generalized to other choices of primitives. We rely on a simple \emph{ad hoc} algorithm~\cite{SchnabelEffientRANSAC} to obtain an initial set of primitives. We expect this algorithm to generate correct primitives but also many false positives. Our problem then becomes the identification of the correct primitives while rejecting the incorrect ones, by searching the subset of primitives that explains the scene point cloud the best.

While the theoretical combinatorics of this search are huge, as they grow exponentially with the number of extracted primitives, the search is structured by some constraints. For example two primitives should not intersect. To tackle this problem, we take inspiration from a recent work on 3D scene understanding~\cite{mcss2021}.  \cite{mcss2021} proposes to rely on the Monte Carlo Tree Search~(MCTS) algorithm to handle a similar combinatorial problem to select objects' 3D models: The MCTS algorithm is probably best known as the algorithm used by AlphaGo~\cite{silver2016alphago}. It is typically used to explore the tree of possible moves in the game Go because it scales particularly well to high combinatorics. \cite{mcss2021} adapts it to 3D models selection by considering a move as the selection of a 3D model for one object, and showed it performs significantly better than the simple hill-climbing algorithm that is sometimes used for similar problems~\cite{ZouGLH19}. Another advantage of this approach is that it does not impose assumptions on the form of the objective function, unlike other approaches based on graphs, for example~\cite{ImaginingTheUnseenSigga14}.

\begin{figure}[t]
 \centering
 \includegraphics[width=0.9\textwidth]{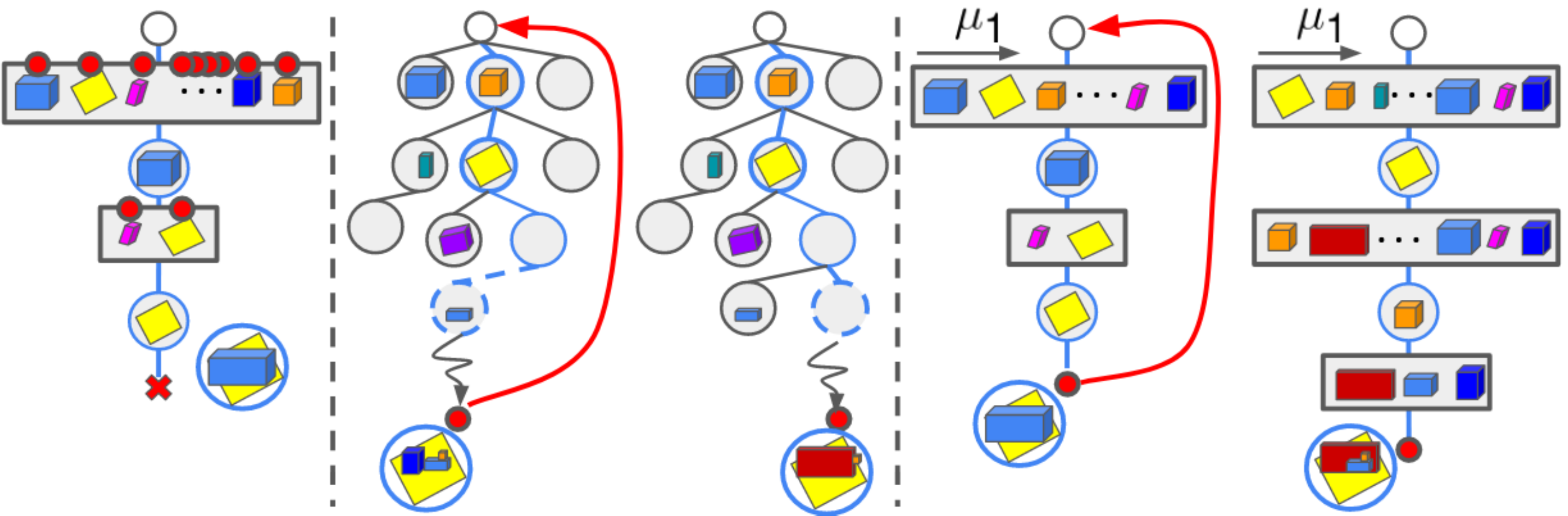}\\
 \flushleft $\>\>\>\>\>\>\>\>$ Hill-Climbing $\>\>\>\>\>\>\>\>$ 2 iterations of MCTS $\>\>\>\>\>\>\>\>$ 2 iterations of our algorithm
 \caption{\textbf{Comparative overview} The hill-climbing algorithm---simply taking the primitive that improves the most the objective function--- can terminate \terminatesymb $\>$ quickly as it gets stuck into a local minimum because of the constraints between primitives. MCTS as used in \cite{mcss2021} explores iteratively the solution tree by traversing \textcolor{blue}{blue paths}, updating which primitives are the most promising ones, but keeping the tree structure fixed. At each iteration, our approach also updates (\updatesymb) which primitives are the most promising ones, and starts with them. This makes our approach identify a good solution much faster than MCTS in general. \textcolor{red}{Red circles} \evalsymb $\>$ represent objective function evaluations. Hill-climbing has to evaluate the complete objective function each time it considers a primitive, while MCTS and our algorithm evaluate the objective function only at the end of an iteration when a complete solution is complete.   }
 \label{fig:insight}
\end{figure}

While exploring the solution tree with MCTS as done in \cite{mcss2021} is efficient, we show we can still speed up the search for a solution significantly more. The tree structure imposes an ordering of the possible 3D models to pick from. Such sequential structures are necessary when MCTS is applied to games as game moves depend on the previous ones, but we argue that there is a more efficient alternative in the case of object detection and selection for scene understanding. 

As illustrated in Figure~\ref{fig:insight}, MCTS works by performing multiple iterations over the tree structure, focusing on the most promising moves. The estimate of how much a move is promising is updated at each iteration. For our problem of primitive selection, we propose to also proceed by iteration. Instead of considering a tree search, at the end of each iteration, we sort the primitives according to how likely they are to belong to the correct solution. The next iteration will thus evaluate a solution that integrates the most promising primitives. Our experiments show that this converges much faster to a correct solution.

To evaluate our approach, we experiment on the ScanNet dataset~\cite{dai2017scannet}, a large and challenging set of indoor 3D RGB-D scans. It contains 3D point clouds of real scenes, with noisy captures and large missing parts, as some parts were not scanned and dark or specular materials are not well captured by the RGB-D cameras.  We did not find any previous work working on similar problems, but we 
adapted other algorithms, 
namely a simple hill-climbing approach~\cite{ZouGLH19} and the MCTS algorithm of \cite{mcss2021} to serve as our baselines for comparison.
To do so, we introduce several metrics to evaluate the fit quality.

Our algorithm is conceptually simple, and can be written in a few lines of pseudo-code. We believe it is much more general than the cuboid fitting problem. It could first be extended to other type of primitives, and applied to many other selection problems with high combinatorics, and could be applied to other 3D scene understanding problems, for auto-labelling for example. We hope it will inspire other researchers for their own problems.

\section{Related Work}
In this section, we first discuss related work on cuboid fitting, and then on possible optimisation methods to solve our selection problem.

\subsection{Cuboid Fitting on Point Clouds}
Primitive fitting is a long standing Computer Vision problem. In the section, we only discuss about methods that operate on point clouds, although there are a large number of methods that are seeking progress in the field of cuboid fitting from 2D RGB images~\cite{roberts1963BlocksWorld,GuptaECCV10,Kluger_2021_CVPR}.

\paragraph{Object scale. }
Sung\etal~\cite{Sung2015} leveraged cuboids decompositions to improve 3D object completion of scans of synthetic objects. Tulsiani\etal~\cite{abstractionTulsiani17} introduced object abstraction using cuboids on more challenging objects from the Shapenet~\cite{shapenet2015} dataset. Paschalidou\etal~\cite{Superquadrics2019CVPR} extended~\cite{abstractionTulsiani17} by using the more expressive superquadrics to fit 3D objects. However these methods only operate at the scale of a single objects, on synthetic data, and always assume or are limited to a moderate number of primitives.
Some older work related to us have focused on parsing an input point cloud as a decomposition into primitives. Li\etal~\cite{li_globFit_sigg11} decompose a \textit{real} scan of an object into primitives by extracting a set of primitives with RANSAC, which they refine by reasoning on relationship between these primitives. However their method works only on very clean scans, and using object that were \textit{built} as a set of primitives. Furthermore, since they reason about interaction between primitives using a graph, the complexity of there method quickly becomes untractable. 

\paragraph{Room-scale cuboid detection.}
Another class of works has focused on room-scale 3D point cloud parsing with cuboids. 
A large number of works focused on detecting object bounding boxes in 3D scans have recently emerged since the deep learning era~\cite{qi2019deep,pvrcnn,shi2020point}. Guo\etal~\cite{guo2020deep} wrote a great survey regarding these methods. Contrary to these methods, our method is able to parse 3D scans with cuboids at the granularity level of parts of objects. Liang\etal~\cite{RGBDcuboid} used RGB-D images to fit cuboids to the point cloud obtained by the depth map. In contrast to us, they operate using single-view images, but also leverage color cues via superpixels.
Shao\etal~\cite{ImaginingTheUnseenSigga14} also parse depth maps with cuboids. Given an initial set of cuboids, they build a graph to exploit physical constraints between them to refine the cuboids arrangement. However, they still require human-in-the-loop for challenging scenes, and their graph based method limits the number of cuboids that can be retrieved without exceeding complexity. Our method, in contrast, can deal with number of cuboids that are an order of magnitude larger.

\subsection{Solution Search for Scene Understanding}

We focus here on scene understanding methods which, like us, do not rely on supervised training data for complete scenes, even if some of them require training data to recognize the objects.  These methods typically start from a set of possible hypotheses for the objects present in the scene (similar to the primitives in our case), and choose the correct ones with some optimization algorithms. 

\textbf{Monte Carlo Markov Chain} (MCMC)~\cite{mcmc} is a popular algorithm to select the correct objects in a scene by imposing constraints on their arrangement. MCMCs can be applied to a parse graph~\cite{Zhao,choi,huang2018holistic,Chen2019HolisticSU} that defines constraints between objects. However, this parse graph needs to be defined manually or learned from manual annotations. Also, MCMCs typically converge very slowly.

\textbf{Greedy approaches}
were also used in previous works
\cite{izadinia2017im2cad}, and they rely on a hill-climbing method to find the objects' poses~\cite{izadinia2017im2cad}. \cite{ZouGLH19} selects objects using hill-climbing as well by starting from the objects with the best fits to an RGB-D image. While simple and greedy, this approach can work well on simple scenes. However, it can easily get stuck on complex situations, as our experiments show. \cite{Lee2010EstimatingSL} uses beam search but this is also an approximation as it also cuts some hypotheses to speed up the search. 

\textbf{Monte Carlo Tree Search} (MCTS) was recently used in \cite{mcss2021}, where they proposed to use MCTS as an optimization algorithm to choose objects that explain an RGB-D sequence. \cite{mcss2021} adapts MCTS 
 by considering the selection of one object as a possible move in a game. The moves are selected to optimize an objective function based on the semantic segmentation of the images and the depth maps. The advantage of this approach is that MCTS can scale to complex scenes, while optimizing a complex objective function. 
 
Our approach is motivated by \cite{mcss2021}. However, we generate the primitives in a very different way, but more importantly, we propose a novel optimization algorithm, which, contrary to MCTS, does not rely on a tree structure, making it is simpler and significantly more efficient than MCTS, as demonstrated by our experiments. 

\begin{table}[t]
    \centering \addtolength{\tabcolsep}{6pt}
    \begin{tabular}{@{}l c c c @{}}
    \toprule
    Method & Uphill & MCTS & Ours\\
    \hline
    Exploratory & \xmark &  \cmark & \cmark \\
    Stochastic & \xmark & \cmark & \cmark \\
    Leverage order invariance & \cmark & \xmark & \cmark \\ 
    \bottomrule
    \end{tabular}
    \caption{\textbf{Properties of different solution search methods.} Our method leverages all popular mechanisms for efficient solution search while leveraging the structure of the problem, which does not require employing tree structures for solution search.}
    \label{tab:procons_algos}
\end{table}

\section{Method}
In this section, we first describe how we extract a large pool of cuboids from a given 3D scan. Then, we formalize the selection of the optimal cuboid arrangement. Finally, we detail the solution we propose.

\begin{figure}[t]
    \centering
    \includegraphics[width=1\textwidth]{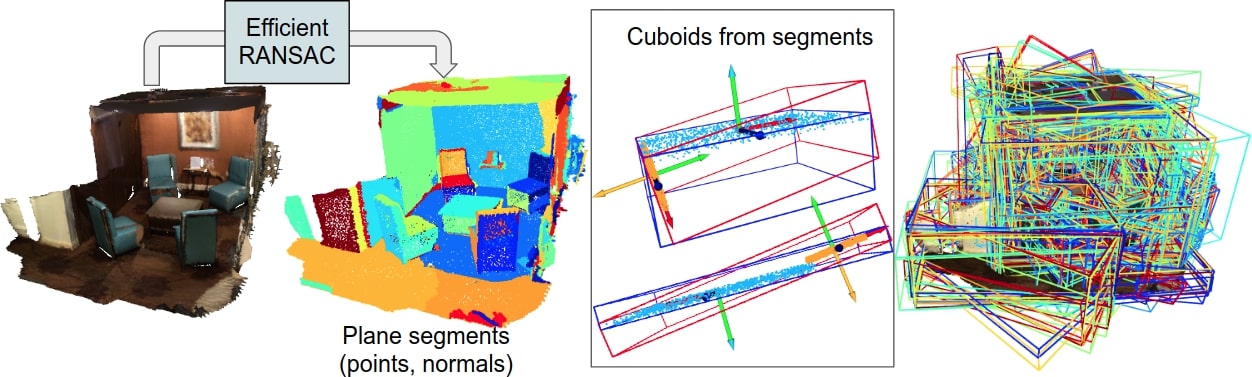}\\
    \caption{\textbf{Overview of our cuboid generation pipeline.} After extracting plane segments using an off-the-shelf algorithm~\cite{SchnabelEffientRANSAC}, we construct cuboids around these segments and pairs of adjacent segments. The result is a dense set of cuboids, which may contain many false positives.}
    \label{fig:box_extraction}
\end{figure}

\subsection{Generating Cuboid Proposals from Noisy Scans}
\label{ssec:box_gen}

Figure~\ref{fig:box_extraction} summarizes our cuboid proposal generation pipeline. 
The goal of this pipeline is to provide a large pool of cuboids. Some extracted cuboids can be false positives at this stage. The correct subset of cuboids will be selected by the next stage.
In this way, we can be robust to noise and missing data in the 3D scan. Our pipeline can be divided in 3 steps: (1) we first extract plane segments; (2) we construct cuboids from pairs of plane segments; (3) we also construct \textit{thin} cuboids by fitting a 3D bounding box to the each plane segment individually. These thin cuboids allow us to represent planar surfaces as well in the final representation.  On average, we obtain 880 cuboids and 174 \textit{thin} cuboids per scene. 

\paragraph{\bf Extracting planes segments. }
We use Efficient-RANSAC by Schnabel~\etal~\cite{SchnabelEffientRANSAC} to extract 3D planes from the input point cloud.  Efficient-RANSAC identifies and returns planar connected components made of 3D points. It is controlled by three hyperparameters: a threshold on the plane-to-point distance to count the inliers, a threshold on the cosine-similarity between normals to points, and a connectivity radius. We use the same hyperparameters for all the scenes in ScanNet, although we could run RANSAC multiple times with various geometric parameters in order to adapt to various types of noise, and still be able to efficiently filter out false positives.

\paragraph{\bf Constructing boxes from pairs of planes.}
Given a set of planes segments $\{\pi_i = (X_i, \bN_i)\}$, where a plane segment $\pi$ is represented as a point cloud $X$ and its fitted plane normal $\bN$, we construct bounding boxes from all pairs of planes ($\pi_A, \pi_B$) that satisfy two criteria, \textit{alignment} and \textit{proximity}. Alignment means that the two normals should be orthogonal or co-linear. Proximity enforces planes segments to have at least one connected component in 3D. We then employ two Gram-Schmidt orthonormalizations to obtain the frame coordinate of two bounding boxes, which are computed to enclose  $X_A \cup X_B$. More details can be found in the supplementary material.

\paragraph{\bf Fitting 3D bounding boxes to 3D plane segments. }
Since we want our method to also retrieve thin objects that may not have compatible neighbors, we therefore fit a 3D oriented bounding box to each plane segment's point cloud $X$, using the efficient ``\textit{Oriented Bounding Box}'' method from~\cite{CGAL_OrientedBB}.\footnote{We used CGAL's~\cite{CGAL_library} implementation of~\cite{SchnabelEffientRANSAC,CGAL_OrientedBB}.}

\subsection{The Cuboids Arrangement Search Problem}

We now want to select a subset $\calS '$ of $\calS$, the set of  cuboids generated in Section~\ref{ssec:box_gen}, which fits well the input point cloud $X$ of the scene. The cuboids in $\calS '$ should not mutually intersect to ensure a minimal representation of this scene.

To solve this problem, we consider (1) an objective function $\ell$, defined in Algorithm~\ref{algo:loss}, which will guide the search towards the best solution, and (2) a search algorithm such as the baselines described in Section~\ref{ssec:search_algos}, that should be designed to converge to the best solution as efficiently as possible. To better present the algorithms, we introduce a \FuncSty{Cuboid} Class, which we present first.

\paragraph{\bf \FuncSty{Cuboid} Class. }
We define a \Cuboid class to instantiate cuboids for our solution search algorithms. It is described by its faces normals and its 8 corners, yielding a surface mesh from which we can sample 3D points. Other attributes can be added to a \Cuboid, depending on the needs of a particular algorithm, e.g. the number of times a \Cuboid $s$ has been used in a solution can be denoted as $s.n_1$. 

To enforce constraints between cuboids, we need to test if the intersection between two cuboids is small enough. We define this criterion using a variation of the measure of a Intersection-over-Union criterion, and provide its pseudo-code in Supp Mat. \FuncSty{isCompatible($s_1, s_2, \eta$)} measures the ratio between the volume $vol(s_1\cap s_2)$ of the intersection between both cuboids $s_1$ and $s_2$, and the minimum of the volumes of each cuboid $vol(s_1)$ and $vol(s_2)$. In practice, we approximate these volumes by uniformly randomly sampling points from both cuboids and count the points that are inside both $s_1$ and $s_2$. The volume ratio is then compared to a threshold $\eta$, to decide if the two \Cuboid intersect. While this test can be performed ``\textit{on the fly}'' when searching solutions, we pre-compute the pair-wise \Cuboid compatibility matrix in advance for efficiency.

\paragraph{\bf Objective Function}
\label{ssec:loss_function}
We aim to minimize the distance between our cuboids and the target point cloud, while keeping its normals aligned with the pointcloud's normals. We use Chamfer Distance (CD) and Cosine Dissimilarity, \emph{i.e.} the complement of Cosine Similarity, as our distance and normals deviation losses, yielding full objective function is described in Algorithm~\ref{algo:loss}. In the loss, we truncate CD to $\tau=0.1$, and normalize it by $\tau$.

\begin{algorithm}[t]
\SetKwInOut{Input}{Input}
\SetKwInOut{Init}{Initialize}
\SetAlgoLined
\SetKwFunction{iscompatible}{\text{isCompatible}}\SetKwFunction{evalloss}{\text{evalObjFunc}}
\SetKwProg{myproc}{procedure}{}{}

\SetKwProg{class}{class}{}{}
\myproc{\evalloss{S, Y}}{
  \Input{Set of cuboids S, target point cloud Y and its normals $\bN(Y)$\;}
  $(X, \bN(X))$ $\gets$ sample\_mesh\_surface(S)\;
  $\loss_c:=$ $\CD(X\rightarrow Y) + \CD(Y\rightarrow X)$\;
  $\loss_n:=$ $\ND(\bN(X)\rightarrow \bN(Y)) + \ND(\bN(Y)\rightarrow \bN(X))$\;
  \KwRet $\loss_c \cdot (1+0.25 \cdot\textrm{exp}(\loss_n))$\;}{}
  \caption{Loss function \evalsymb}
 \label{algo:loss}
\end{algorithm}

\subsection{Solution Search Baseline Algorithms}
\label{ssec:search_algos}

\subsubsection{Hill-Climbing Algorithm.}
\label{ssec:uphill}
The first baseline for our discrete optimization problem is the Hill-Climbing algorithm~\cite{hillclimbing}, a naive greedy descent algorithm. This algorithm constructs a solution iteratively, where at each iteration, it comprehensively searches for the proposal that best improves the loss function of a solution $\calS_F$, while leaving the solution valid \textit{i.e.} with no incompatibilities. If no proposal is available nor can improve the objective function, the algorithm stops \terminatesymb. The pseudo code for Hill-Climbing is given in the supplementary material.

\subsubsection{MCTS Algorithm.}
\label{sec:MCTS}
We first describe here the MCTS algorithm, as it inspired our algorithm. \cite{browne2012survey} provides a full description of the MCTS algorithm. We present it in the context of our cuboid selection problem, following what was done in \cite{mcss2021} for 3D model selection. \cite{mcss2021} provides a pseudo code for MCTS.

MCTS is able to efficiently explore the large trees that result from the high combinatorics of some games such as Go. As represented in Figure~\ref{fig:insight}, the nodes of the tree correspond to possible states, and the branches to possible moves. MCTS does not build explicitly the entire tree---this would not be tractable anyway---, but only a portion of it, starting from the root at the top.

\paragraph{\simulatesymb Simulation step.} Nodes are thus created progressively at each iteration. To decide which nodes should be created, the existing nodes contain in addition to a state an estimate $V$ of the \emph{value} of this state. To initialize $V$, MCTS uses a \emph{simulation step} denoted $\boldsymbol{\rightsquigarrow}$ in Figure~\ref{fig:insight}, which explores randomly the rest of the tree until reaching a leaf without having to build the tree explicitly. For games, reaching a leaf corresponds to either winning or loosing the game. If the game is won, $V$ should be large; if the game is lost, $V$ should be small. 

\paragraph{Adaptation to our problem.} Figure~\ref{fig:insight} shows that in our case, a state in a node is the set of primitives that have been selected so far. A ``move''  corresponds to adding a primitive to the selected primitives. The children of a node contain primitives that are mutually incompatible, and compatible with the primitives in the ancestor nodes: Such structure ensures that every path in the tree represents a valid solution. In this paper, we consider two possibilities: A varying number of children as in \cite{mcss2021} and MCTS-Binary, a binary tree version of MCTS: In MCTS-Binary, a node has two children, corresponding to selecting or skipping a primitive. More details are provided in the supplementary material.

\terminatesymb ``Reaching a leaf'' happens when no more primitives can be added, because we ran out of primitives or because all the remaining primitives intersect with the primitives already selected. The value $V$ of the new nodes are initialized after the simulation step by evaluating the objective function \evalsymb
for the set of primitives for the leaf. We take this objective function as a fitness measure between the primitives and the point cloud. Note that this function does not need to have special properties, nor do we need heuristics to guide the tree search.

\newtvar{crnt}
\newtvar{UCB}
\newtvar{Children}

\paragraph{Selection and expansion steps.} 
At each iteration, MCTS traverses the tree starting from the root node, often using the standard Upper Confidence Bound~(UCB) criterion~\cite{browne2012survey} to choose which branch to follow. A high UCB score for a node means that it is more likely to be part of the correct solution. This criterion depends on the values $V$ stored in the nodes and balances exploitation and exploration: When at a node $N$, we continue with its child node $N'$ that maximizes the UCB score, which depends on the number of times $N$ and $N'$ have been visited so far. This criterion allows MCTS to balance exploration and exploitation.

At some point of this traversal procedure, we will encounter a node with a child node $N$ that has not been created yet, we add the child node to the tree. We use the simulation step described above to initialize $V(N)$ and initialize $n(N)$ to 1.

\paragraph{\updatesymb Update step.}   MCTS also uses the value $V(N)$ to improve the value estimate of each node $N'$ visited during the tree traversal. Different ways to do so are possible, and we found that for our problem, it is better to take the maximum between the current estimate $V(N')$ and $V(N)$: $V(N') \leftarrow \max(V(N'), V(N))$. $n(N')$, the number of times the node was visited is also incremented.

\paragraph{Final solution.} After a chosen number of iterations, MCTS stops. For our problem, we obtain a set of primitives by doing a tree traversal starting from the root node and following the nodes with the highest values $V$.

\subsection{Our algorithm: MonteBoxFinder}

We first review the issues when using MCTS for our problem, then give an overview of our algorithm and its components. Finally, we provide some details for each component.

\subsubsection{Moving from MCTS.}
\label{sec:mcts2ours}
Our primitives selection algorithm is inspired by MCTS, and it is motivated by two observations that show that MCTS is not optimal for our selection problem:
\begin{itemize}
    \item the order we select the primitives does not matter. However, MCTS keeps growing its tree without modifying the nodes already created. This implies that if a primitive appears at the top of the tree but does not actually belong to the correct solution, it will slow down the convergence of MCTS towards this solution.
    \item if a node corresponding to adding some primitive $P$ has a high value $V$, the node corresponding to not keeping $P$ should have a low value, and vice versa. There is no mechanism in MCTS as used in \cite{mcss} to ensure this. This is unfortunate as one iteration could be used to update more nodes than only the visited nodes.
\end{itemize}


\begin{algorithm}[t]
\begin{multicols}{2}
\SetKwInOut{Input}{Input}
\SetKwProg{Init}{Initialize}{}{}
\SetAlgoLined
\SetKwFunction{updateMC}{\text{UpdateMC}}
\SetKwFunction{updateMBF}{\text{Update}}
\SetKwFunction{initNodes}{\text{InitializeNodes}}
\SetKwFunction{simulateMC}{\text{SimulateMC}}
\SetKwFunction{simulateMBF}{\text{Simulate}}
\SetKwProg{myproc}{procedure}{}{}
\KwResult{Set of selected \Cuboid $\calS_F$}
\Input{Set of available \Cuboid $\calS$\;}
 Number of evaluations $N_{eval}$ \;
 Threshold $\eta$ \;
 Current solution $\calS_c := \emptyset$ \;
 Final solution $\calS_F := \emptyset$\;
 Current best loss $\loss^* := +\infty$ \;
\myproc{\initNodes{$\calS$}}{
  \Input{Pool of \Cuboid $\calS$\;}
    $\calS \gets \textrm{Shuffle}(s \in \calS)$\;
    $\calS_c \gets $\simulateMBF{$\calS$, $\eta$}\;
    $\loss \gets$ \evalloss{$\calS_c$}\;
    \tcp{Update ALL \Cuboid states}
    $\calS \gets $ \updateMBF{$\calS$, $\calS_c$, $\loss$}\;
 \KwRet $\calS$
 }{}
 ~\\
 ~\\ 
 
\tcp{MonteBoxFinder Core Algorithm}
$\calS \gets $ \initNodes{$\calS$}\;
\For{iter=0; iter$ \neq N_{eval}$; iter++}{
$\calS \gets \textrm{Sorted}_\downarrow(s \in \calS, s \mapsto s.\mu_1)$\;
$\calS_c \gets $\simulateMBF{$\calS$, $\eta$}\;
$\loss \gets$ \evalloss{$\calS_c$}\;
\tcp{Update ALL \Cuboid states}
$\calS \gets $ \updateMBF{$\calS$, $\calS_c$, $\loss$}\;
\uIf{$\loss < \loss^*$}{
    $\loss^* \gets \loss$\;
    $\calS_F \gets \calS_c$\;
}
}
\KwRet{Best solution $\calS_F$}\;
~\\
~\\
\end{multicols}
 \caption{Our MonteBoxFinder Algorithm}
 \label{algo:MBF}
\end{algorithm}

\subsubsection{Overview.}

We give an overview of our algorithm in Algorithm~\ref{algo:MBF}. To exploit the two observations described above, we do not use a tree structure. Instead, we use the list of primitives which we sort at each iteration, 
by exploiting our current estimate for each primitive to be part of the current solution. 
Our method progressively estimates and exploits a prior probability $\calP$ for a primitive to belong to the solution based on our adaptation of the Upper Bounding Criterion~(UCB) that balances the exploitation vs. exploration trade-off.

\begin{figure*}
\noindent
\centering
\begin{algorithm}[H]
\begin{multicols}{2}
\SetKwInOut{Input}{Input}
\SetKwProg{Init}{Initialize}{}{}
\SetAlgoLined
\SetKwFunction{update}{\text{Update}}
\SetKwFunction{simulate}{\text{Simulate}}
\SetKwProg{myproc}{procedure}{}{}
\Input{Exploration probability $\calP_\epsilon$\;} 
Threshold $\delta$\;
\myproc{\simulate{$\calS_A$, $\eta$}}{
  \Input{Pool of available \Cuboid $\calS_A$, 
         threshold $\eta$}
  Output $\calS_F := \emptyset$\;
  \For{$s \in \calS_A$}{ 
    \uIf{$s.$\iscompatible{$\calS_F$, $\eta$}}{
    $\epsilon := \textrm{uniform\_sample}([0,1])$\;
        \uIf{ $(\epsilon < \calP_\epsilon)$}{ 
            \uIf{$(s.\mu_1 > s.\mu_0)$}{
                $\calS_F$.add($s$)
            }
        } \Else{ 
            \uIf{$(s.\mu_1 < s.\mu_0)$}{
            $\calS_F$.add($s$)
            }
        }
    }
  }
  \KwRet $\calS_F$}{}
  
\myproc{\update{$\calS$, $\calS_F$, $\loss$}}{
  \Input{Full pool of \Cuboid $\calS$\; 
  Selected set of \Cuboid $\calS_F\subset \calS$\; Solution score $\loss$\;}
  \For{$s \in \calS$}{ 
    \uIf{$s \in \calS_F$}{
        \tcp{Update best $\loss$ when kept}
        $s.\loss_1$ $\gets \min(\loss, $ $s.\loss_1)$ \\
        $s.n_1 \gets s.n_1 + 1$ \\
        $s.\mu_1 \gets -s.\loss_1 + \sqrt{\ln(1 / \delta) / s.n_1}$\\
    } \Else{
        \tcp{Update best $\loss$ when rejected}
        $s.\loss_0$ $\gets \min(\loss, $ $s.\loss_0)$ \\
        $s.n_0 \gets s.n_0 + 1$ \\
        $s.\mu_0 \gets -s.\loss_0 + \sqrt{\ln(1 / \delta) / s.n_0}$\\
    }
  }
  \KwRet $\calS$}{}
  ~ \\
\end{multicols}
\caption{Simulate(\simulatesymb) and Update(\updatesymb) functions of our algorithm}
 \label{algo:UCB}
\end{algorithm}
\end{figure*}

\subsubsection{Initialization.}
We initialize the run with a few random traversals in order to initialize the states of each \Cuboid proposal.

\subsubsection{Simulate. (\simulatesymb)}
At every iteration we first sort primitives $\calS_A$ according to their confidence value $s.\mu_1$ in descending order, hence more confident primitives will be more likely selected. Afterwards, we perform the simulation that pops primitives $s$ from sorted $\calS_A$. With probability $\calP_\epsilon=0.3$, we perform exploitation and add $s$ to the list of selected proposals $\calS_F$ if $(s.\mu_1 > s.\mu_0)$. Otherwise, we perform exploration and add $s$ to $\calS_F$ if $(s.\mu_1 < s.\mu_0)$.

\subsubsection{UCB Criterion.} We modified the UCB score to fit our algorithm, which does not rely on a tree structure. We use this modified term to estimate two confidence measures $s.\mu_0$ and $s.\mu_1$ reflecting how much a cuboid $s$ is likely to belong to the correct solution or not:
 \begin{equation}
  s.\mu_0 = -s.\loss_0 + \sqrt{\ln(1 / \delta) / s.n_0} , \quad
  s.\mu_1 = -s.\loss_1 + \sqrt{\ln(1 / \delta) / s.n_1} ,
  \label{eq:UCBour}
    \end{equation}
where $s.\rho_0$ and $s.\rho_1$ are the minimum loss values reached when rejecting and accepting primitive $s$, $s.n_0$ and $s.n_1$ denote the number of times that the primitive were rejected and selected respectively, and $\delta=0.03$ is a hyperparameter modifying the exploration rate, smaller $\delta$ implies larger exploration.

\subsubsection{Update (\updatesymb)} In comparison with the \textit{update step} of MCTS described in \ref{ssec:search_algos}, our MonteBoxFinder algorithm updates \textit{all} primitives states after an iteration. If a primitive $s$ was selected, we update its $s.\loss_1$, $s.\mu_1$, and $s.n_1$ values based on the obtained loss $\loss$ and our adapted UCB criterion, otherwise we update its $s.\loss_0$, $s.n_0$, and $s.\mu_0$ values instead. In the next iteration during simulation, we use these value to determine whether to select or reject the primitive.

\section{Experiments}
\subsection{Dataset}

ScanNet~\cite{dai2017scannet} is a dataset that contains noisy 3D scans of 1613 indoor scenes. We evaluate our method on the full dataset, where for each scene, we used the decimated and cleaned point clouds provided in \cite{dai2017scannet} both for the box proposals generation step and for the solution search step.

\subsection{Metrics}
\subsubsection{Fitness measures. }
The most direct way to measure the quality of a solution is to measure the loss function $\ell$ described in Algorithm~\ref{algo:loss}. Indeed, we want to evaluate the ability of our algorithm to search the solution space. Additionally, we measure a bi-directional precision metric \accuracy. \accuracy is computed as the  proportion of points successfully matched between the ``\textit{synthetic}'' point cloud $X$, generated by sampling 3D points from retrieved 3D cuboid meshes, and the 3D scan $Y$. A point is successfully matched if its Chamfer Distance (CD)\footnote{In this case, we do not apply the normalization discussed in Algorithm \ref{algo:loss}} value is below a threshold $\tau=0.2$:
\begin{equation}
    \text{\accuracy} = \dfrac{|\{x\in X \>\> \textrm{s.t.} \>\> CD[x\rightarrow Y] \leq \tau\}|}{2|X|} +  \dfrac{|\{y\in Y \>\> \textrm{s.t.} \>\> CD[y\rightarrow X] \leq \tau\}|}{2|Y|} \> .
\end{equation}

\subsubsection{Efficiency measure. }
\begin{figure}[t]
    \centering
    \includegraphics[width=0.9\linewidth]{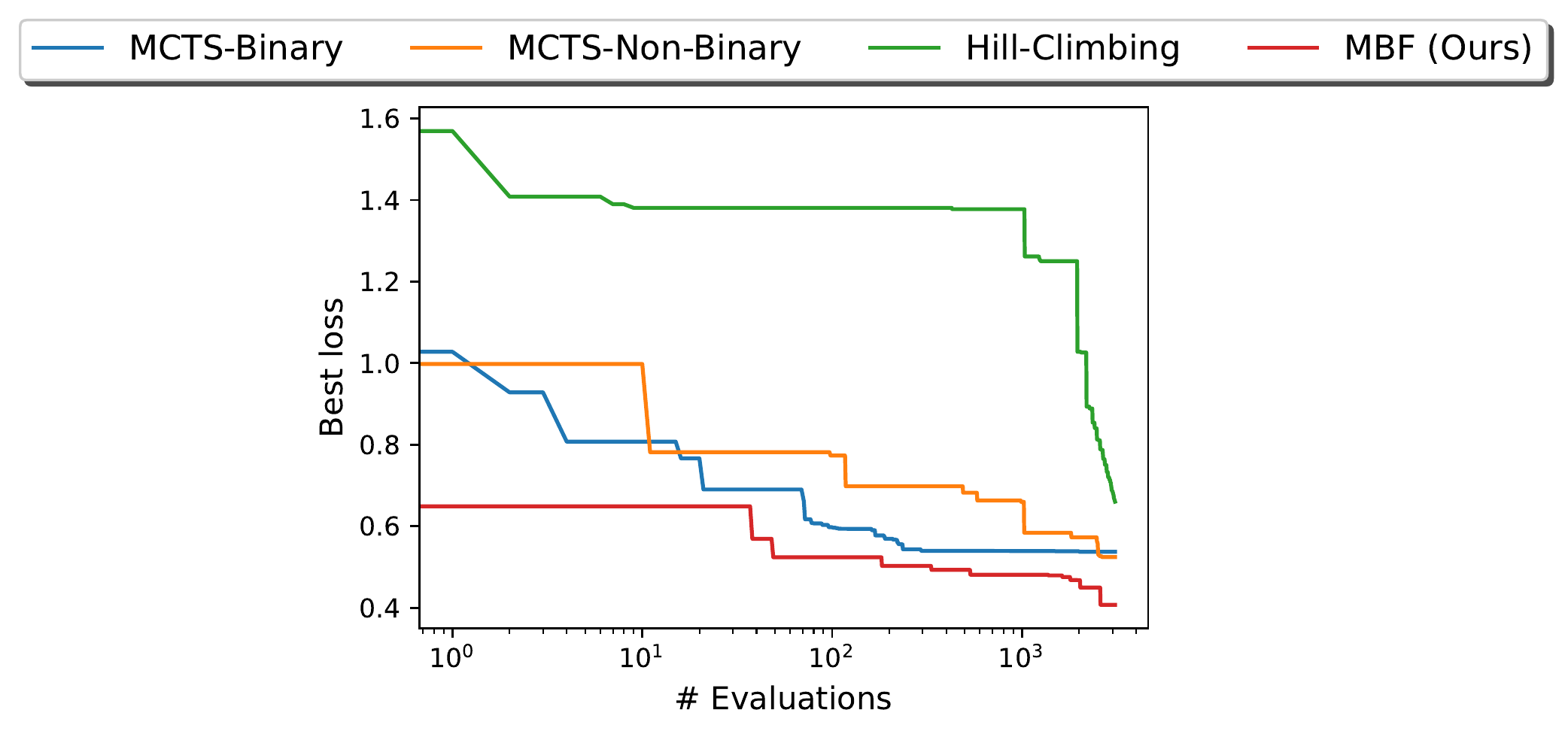}
    \caption{{\bf Value of the objective function for the best found solution as a function of the number of evaluations for Hill-Climbing, MCTS, MCTS-Binary, and our MonteBoxFinder~(MBF) method.} Hill-Climbing requires many evaluations before finding a reasonable solution, which explains the flat curve at the beginning. It also gets stuck into a local minimum and stops improving. In this experiment, we give the number of evaluations Hill-Climbing used before getting stuck to the three other methods. Our method converges significantly faster than the other methods towards a better solution. Similar graphs for other scenes are provided in the supplementary. }
    \label{fig:progress_comparison}
\end{figure}

The motivation for developing our approach compared to \cite{mcss2021} is to converge faster  towards a good solution. In order to measure efficiency of a given method,
we consider the curve of the objective function of the best found solution as a function of the iteration, as the ones showed in Figure~\ref{fig:progress_comparison}. We use the 
 Area Under the Curve (AUC) given a maximum budget of iterations $N_\text{eval}$: the lower the AUC, the faster the convergence. We also report AUC~(norm), which normalizes the AUC values of the different between $0$ and $1$, with $0$ being the value of the best performing method and $1$ being the value of the worst performing method.

\subsubsection{Complexity measure. } 
We observe that bad solutions tend to contain a small number of selected primitives. This is because it is challenging to find a large  subset of cuboids with no intersection between any pair of cuboids. Hence we also report the number of cuboids in the retrieved solutions.

\subsection{Evaluation Protocol}
\label{sec:eval_protocol}
For all scenes from the ScanNet dataset~\cite{dai2017scannet}, we run the Hill-Climbing method, and obtain its solution $\calS_\text{HC}$. We then consider the number $N_\text{eval}$ of evaluations of the objective that were required by Hill-Climbing to construct this solution. We then run MCTS and our algorithm using the same number of evaluations $N_\text{eval}$. This ensures the three methods are compared fairly, as they are given the same evaluation budget, which is by far the most costly step of all three algorithms.

\subsection{Quantitative results}
\label{sec:quant}
\newcommand{\numbf}[1]{\bf #1}
\newcommand{\mrt}{1pt}

\begin{table}[t]
    \centering \addtolength{\tabcolsep}{6pt}
    \begin{tabular}{@{}l
    		>{\centering\arraybackslash}m{0.04\textwidth}
    		>{\centering\arraybackslash}m{0.14\textwidth}
    		>{\centering\arraybackslash}m{0.1\textwidth}
    		>{\centering\arraybackslash}m{0.07\textwidth}
    		>{\centering\arraybackslash}m{0.14\textwidth}@{}}
    \toprule[\mrt]
    & Loss$\downarrow$ & Precision $\uparrow$ & AUC $\downarrow$ & AUC (norm)~$\downarrow$ & Avg. \# Cuboids~$\uparrow$ \\
    \midrule
     Hill-Climbing & 0.383 & 0.928 & 0.871 & 0.998 & 12\\
     MCTS & 0.247 & 0.966 & 0.427 & 0.225 & 28\\
     MCTS-Binary & 0.292 & 0.961 & 0.370 & 0.102 & 35\\
     Ours~(MonteBoxFinder) & \textbf{0.201} & \textbf{0.982} & \textbf{0.322} & \textbf{0.018} & \textbf{37}\\

    \bottomrule[\mrt]

    \end{tabular}

    \caption{\textbf{Comparison between our method and our baselines.} Our method outperforms all baselines on all metrics computed on ScanNet. We retrieve a more accurate fit, while being able to find more non-intersecting cuboids.    
    }
    \label{tab:benchmark}
\end{table}

Table~\ref{tab:benchmark} provides the results of our experimental comparisons.
As expected, the Hill-Climbing algorithm performs worst: By greedily selecting proposals that minimize the loss, it gets stuck to local minimum solutions consisting of large proposals. It can also provide a complete solution only once it converged, while MCTS, MCTS-Binary and our method can provide a good solution much faster. 
The table also shows that our algorithm converges significantly faster than MCTS and MCTS-Binary, which was the desired goal. 
Interestingly, MCTS-Binary performs better than the original MCTS method of \cite{mcss2021}. In the supplementary material, we discuss in details the links between our method and MCTS-Binary.

\subsection{Qualitative Results}

Figure~\ref{fig:qual} shows qualitative results. Hill-climbing focuses on large cuboids to describe the scene. MCTS often selects many true positives but misses some of the proposals because it cannot explore deeper levels of the tree for the given iteration budget. In contrast, our algorithm is able to successfully retrieve cuboid primitives for objects of different sizes, such as walls, floors, and furniture. 

\newlength{\niceresultwidth}
\setlength{\niceresultwidth}{0.28\linewidth}
\newcommand{\niceresult}[5]{
\includegraphics[width=\niceresultwidth, trim= #2 #3 #4 #5, clip]{images/qual/#1_uh.jpg} & 
\includegraphics[width=\niceresultwidth, trim= #2 #3 #4 #5, clip]{images/qual/#1_mcts.jpg} &
\includegraphics[width=\niceresultwidth, trim= #2 #3 #4 #5, clip]{images/qual/#1_ucb.jpg} \\[-0.8em]}

\begin{figure}[t]
    \centering
    \begin{tabular}{ccc}
    \niceresult{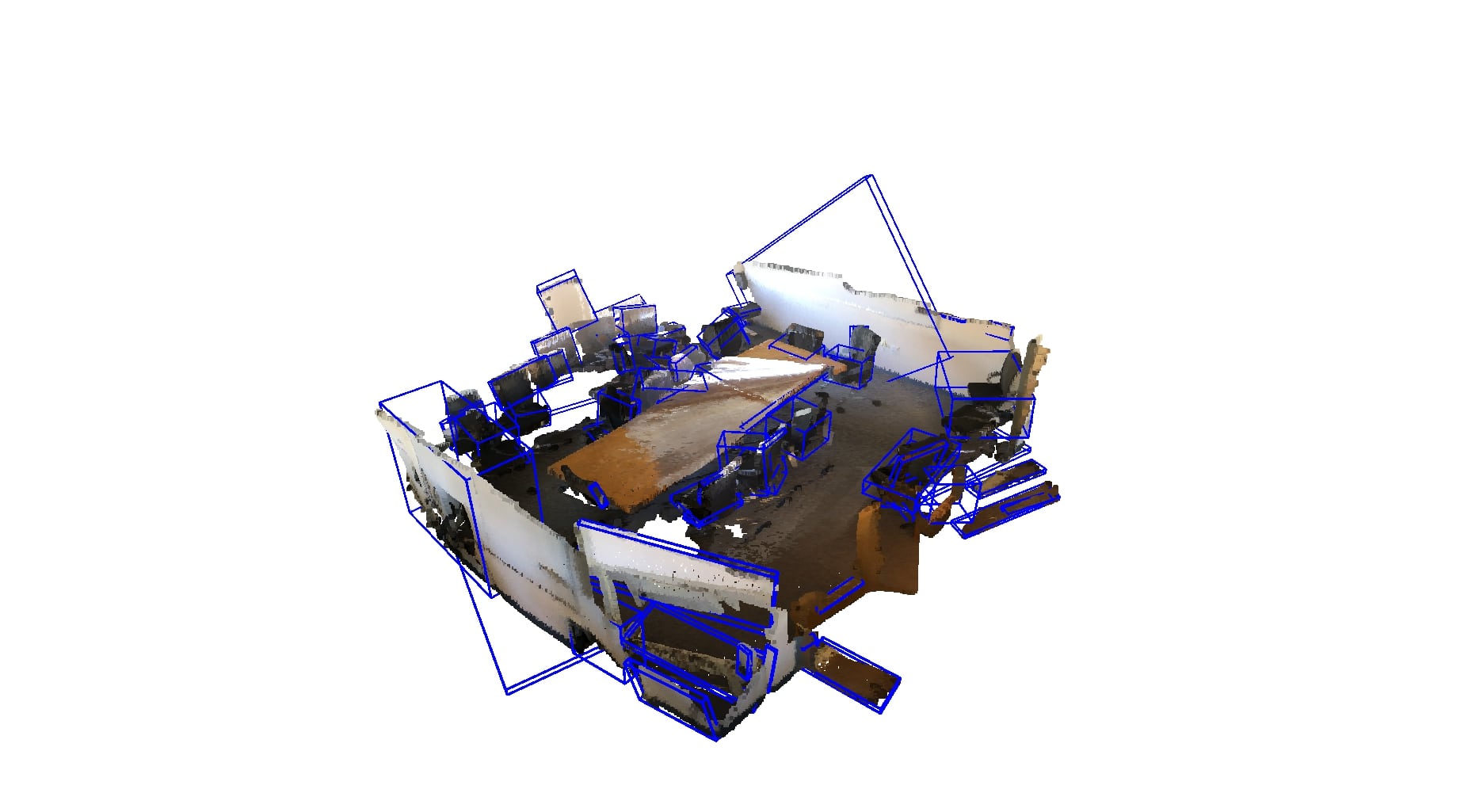}{500}{100}{500}{330}
    \niceresult{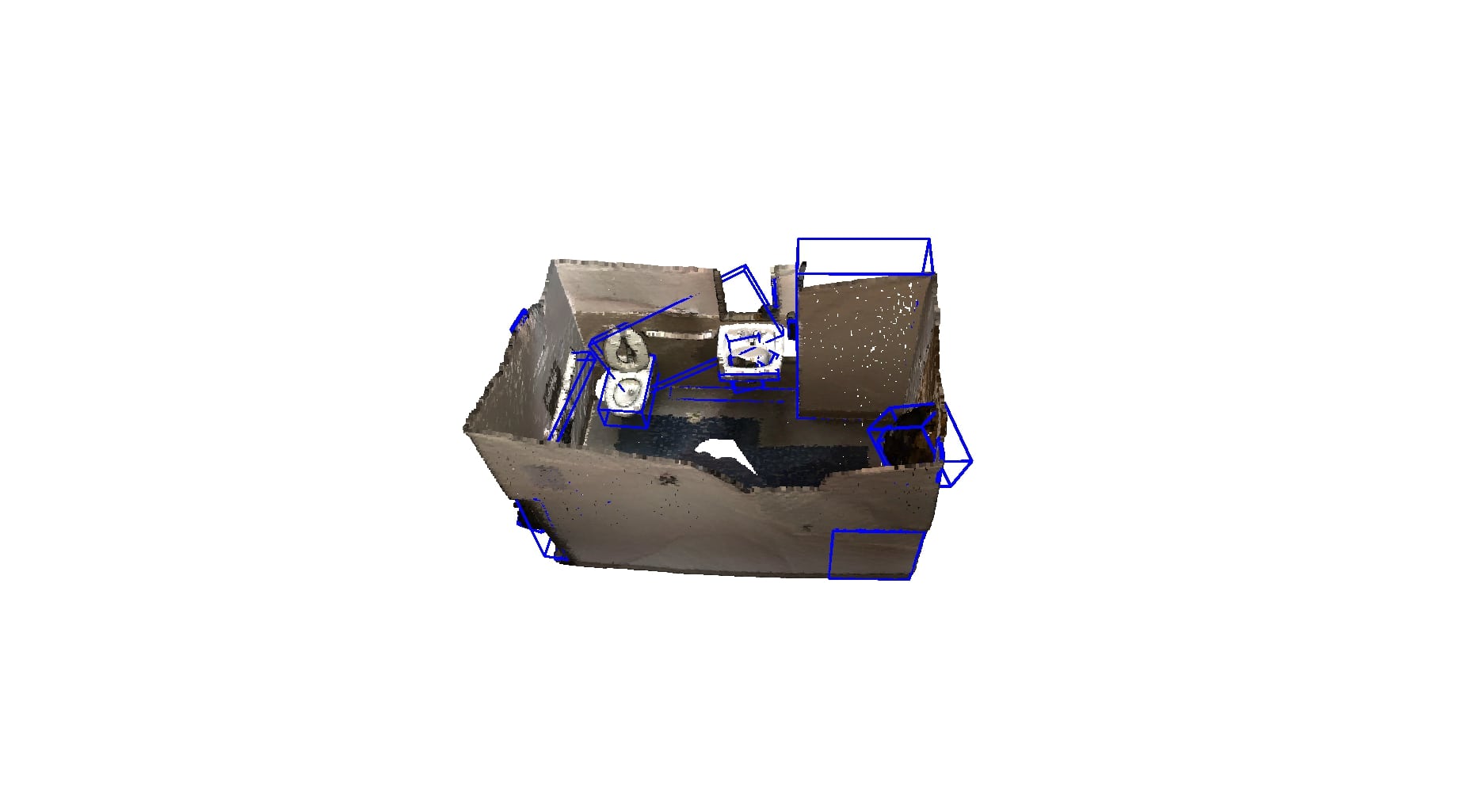}{500}{270}{500}{300}
    \niceresult{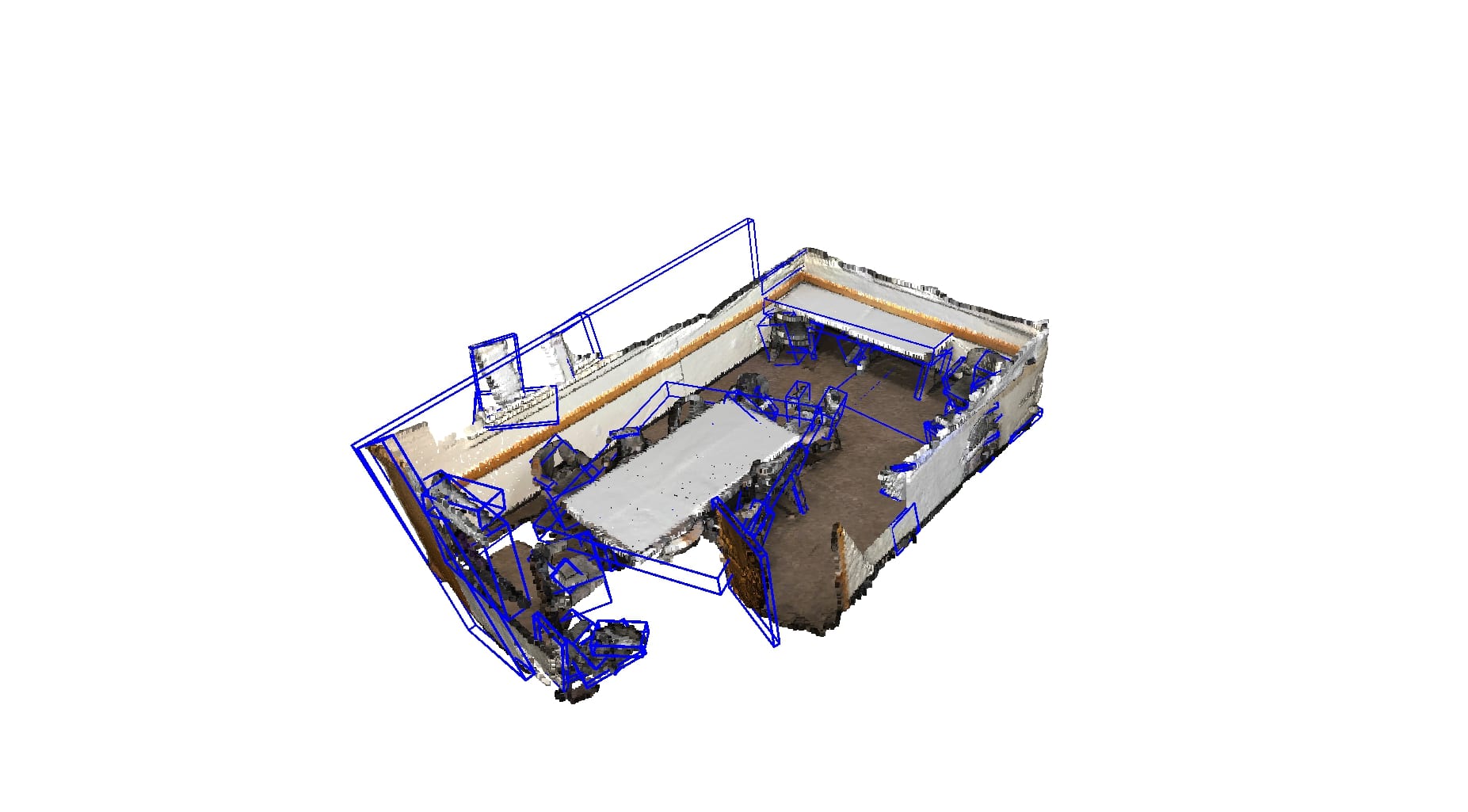}{500}{100}{500}{320}
    \niceresult{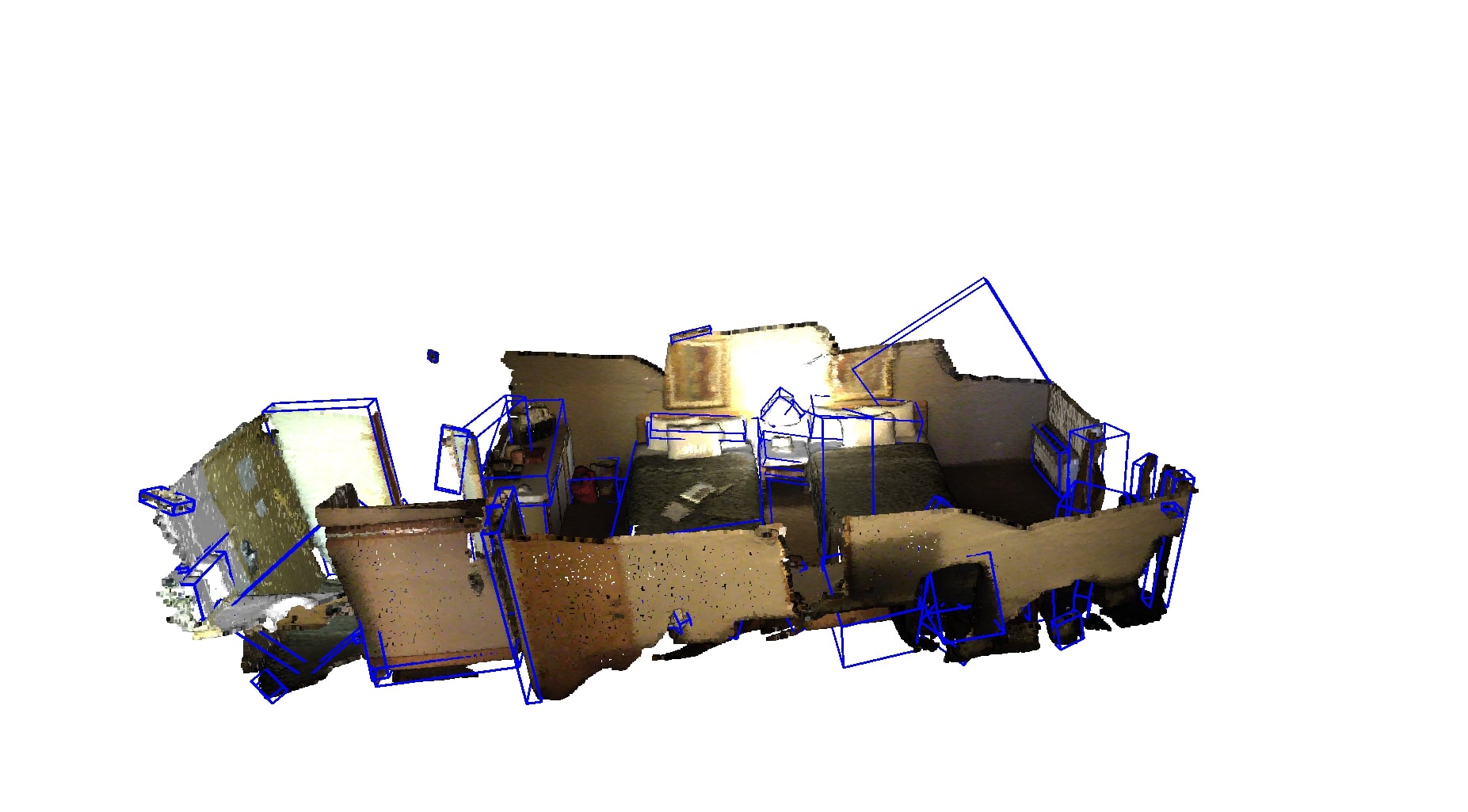}{500}{100}{500}{400}
    \niceresult{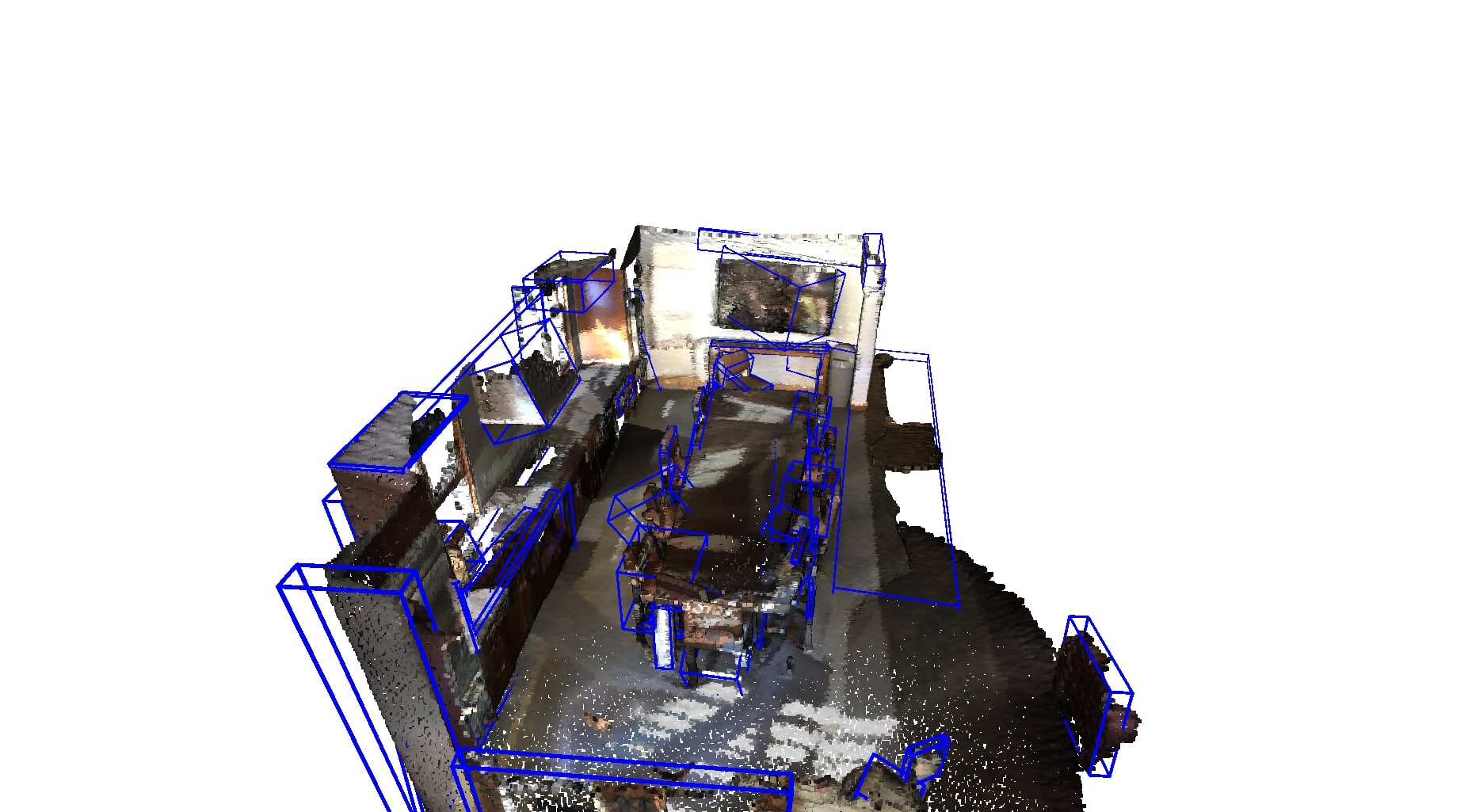}{500}{100}{500}{370}\\[-.6em]
    Hill-climbing & MCTS-Binary & MonteBoxFinder (ours) \\
    \end{tabular}
    \caption{\textbf{Qualitative results.} Hill-climbing often selects large cuboids that span across multiple different objects~(first, third, fourth rows, and fifth rows). MCTS does better, but does not sufficiently explore the solution space~(second row). In contrast, our algorithm outperforms both methods and is able to successfully reconstruct many chairs in first, third, and fifth rows, and bedroom furniture in fourth row. \emph{More qualitative results are provided in the supplementary material.}}
    \label{fig:qual}
\end{figure}

\section{Conclusion}
We proposed a method for efficiently and robustly finding a set of cuboids that fits well a 3D point cloud, even under noise and missing data. Our algorithm is not restricted to cuboids, and could consider other primitives. Only a procedure to identify the primitives is required, even if it generates many false positives as our algorithm can reject them.  Moreover, the output of our algorithm could be used to generate labeled data for training a deep architecture for fast inference. This could be done to predict cuboids from point clouds, but also from RGB-D images, since the 3D scans of ScanNet were created from RGB-D images By simply reprojecting the cuboids retrieved by our method, we can obtain RGB-D images annotated with the visible cuboids.\newline
{\bfseries\noindent Acknowledgments}  We would like to thank Pierre-Alain Langlois for his suggestions and help with CGAL. We thank Gul Varol, Van Nguyen Nguyen and Georgy Ponimatkin for our helpful discussions. This project has received funding from the CHISTERA IPALM project.

%
%
\bibliographystyle{splncs04}
\bibliography{egbib}

\title{MonteBoxFinder: Detecting and Filtering Primitives to Fit a Noisy Point Cloud \\ Supplementary Material}
\titlerunning{MonteBoxFinder}
%
\author{Micha\"el Ramamonjisoa\inst{1} \and
	Sinisa Stekovic\inst{2} \and
	Vincent Lepetit\inst{1,2}}
\authorrunning{M. Ramamonjisoa et al.}
%
\institute{LIGM, Ecole des Ponts, Univ Gustave Eiffel, CNRS, Marne-la-vall\'ee , France \email{first.lastname@enpc.fr}\\ \and
	Institute for Computer Graphics and Vision, Graz University of Technology, Graz,
	Austria
	\email{sinisa.stekovic@icg.tugraz.at}}

\maketitle
\thispagestyle{empty}
\vspace{-10pt}

\section{More on Cuboids}

\subsection{Constructing Cuboids from Pairs of Plane Segments}

In this section, we provide more details regarding the construction of cuboids given a pair of plane segments $\pi_A = (X_A, \bN_A)$ and $\pi_B = (X_B, \bN_B)$. 

\subsubsection{Checking planes adjacency}
We recall that two planes $(\pi_A, \pi_B)$ should be considered for constructing a cuboid only if they fulfill two requirements: \textit{alignment} and \textit{proximity}.
\paragraph{Proximity} Two plane segments are adjacent if they verify the \textit{proximity} criterion, which requires that they have at least one connected component, such that $\gamma > \min\limits_{(x_a, x_b) \in X_A \times X_B} (\|x_a-x_b\|^2_2)$, where $\gamma$ is small.
\paragraph{Alignment} Two plane segments $\pi_A = (X_A, \bN_A)$ and $\pi_B = (X_B, \bN_B)$ are aligned if they are either ``orthogonal enough'' or ``co-linear enough''; This corresponds to $|\bN_A^T \bN_B| < \alpha$ or $|\bN_A^T \bN_B| > \beta$, respectively, where $\alpha << 1$ and $\beta \lesssim 1$.

\noindent In our experiments we set $(\alpha, \beta, \gamma) = (0.3, 0.7, 0.025)$.

\vspace{-10pt}

\subsubsection{Getting cuboids main axes}

We can now construct \textit{two} orthonormal bases $\calB_A = (\bu_A, \bv_A, \bw_A)$ and $\calB_B = (\bu_B, \bv_B, \bw_B)$ of vectors using Gram-Schmidt orthonormalization with $N_A$ or $N_B$ as the first vector alternatively, as shown in Equations~\eqref{eq:uvw_A} and \eqref{eq:uvw_B}.

\begin{subequations}
\setlength{\abovedisplayskip}{0pt}
\setlength{\belowdisplayskip}{0pt}
\setlength{\columnsep}{4em}
\begin{multicols}{2}
\begin{equation}
    \begin{aligned}
    \bu_A &:= \dfrac{\bN_A}{\|\bN_A\|_2} \\
    \bv_A &:= \bN_B - \dfrac{\bu_A^T \bN_B}{\|\bN_B\|_2}\bu_A \\
    \bv_A &\gets \dfrac{\bv_A}{\|\bv_A\|_2}\\
    \bw_A &:= \bu_A \times \bv_A,
    \end{aligned}
    \label{eq:uvw_A}
\end{equation}
    
\begin{equation}
    \begin{aligned}
    \bu_B &:= \dfrac{\bN_B}{\|\bN_B\|_2} \\
    \bv_B &:= \bN_A - \dfrac{\bu_B^T \bN_A}{\|\bN_A\|_2}\bu_B \\
    \bv_B &\gets \dfrac{\bv_B}{\|\bv_B\|_2}\\
    \bw_B &:= \bu_B \times \bv_B,
    \end{aligned}
    \label{eq:uvw_B}
\end{equation}
\end{multicols}
\end{subequations}

\subsubsection{Computing the final cuboids}

Based on the two cuboids bases, we compute their sizes by simply projecting all 3D points $X\in X_A\cup X_B$ and computing the minimum and maximum of the projections along each axes of $\calB_A$ and $\calB_B$. This results in two bounding boxes aligned with $\calB_A$ and $\calB_B$ respectively, which both enclose all points in $X_A \cup X_B$.

\subsection{Computing Intersections with \FuncSty{isCompatible}}

In order to check compatibility between cuboids $s_1$ and $s_2$, we design a variation of an Intersection-over-Union criterion, replacing the \textit{Union} with the volume of the smallest cuboid between $s_1$ and $s_2$. In order to compute the volume of the intersection, we approximate volumes by sampling points in both $s_1$ and $s_2$ and counting points that are inside both. Full details of the procedure used in \FuncSty{isCompatible} are given in Algorithm~\ref{algo:box_intersect}. In practice we use an intersection threshold $\eta=10\%$.

\begin{algorithm}[t]
\SetKwInOut{Input}{Input}
\SetKwInOut{Params}{Geometry}
\SetKwInOut{Init}{Initialize}
\SetAlgoLined
\SetKwFunction{iscompatible}{\text{isCompatible}}\SetKwFunction{iov}{\text{IntersectionOverVolume}}
\SetKwFunction{cuboid}{\text{Cuboid}}
\SetKwProg{myproc}{procedure}{}{}
\myproc{\iscompatible{$s_1$, $\calS$, $\eta$}}{
  \KwResult{Returns True if \Cuboid $s_1$ is compatible with all \Cuboid in $\calS$ }
  \Input{Cuboids  \Cuboid $\calS$, threshold $\eta$}
  \uIf{$(\forall s_2\in\calS, \>$ \iov{$s_1$,$s_2$}$> \eta)$}{
    \KwRet False \;
  }
 \KwRet True \;
}{}
\myproc{\iov{$s_1$, $s_2$}}{
    \Input{Cuboids $s_1$ and $s_2$}
    Volume of \Cuboid $s_1$ $V_1$ $:=$ Volume($s_1$)\\
    Volume of \Cuboid $s_1$ $V_2$ $:=$ Volume($s_2$)\\
    Number of samples $N_{samples} := 5000$\;
    Number of samples from $s_1$ in $s_2$ $N_{1\subset2} := 0$\;
    Number of samples from $s_2$ in $s_1$ $N_{2\subset1} := 0$\;
    
    \tcp{Sample 3D points within both cuboids $s_1$ and $s_2$}
    $\calX_1$ := sample\_points\_inside($s_1$, N) \;
    $\calX_2$ := sample\_points\_inside($s_2$, N) \;
    
    \tcp{Count points sampled in $s_1$ which are also inside $s_2$}
    \For{$x \in \calX_1$}{ 
        \uIf{$x \in s_2$}{
            $N_{1\subset2} \gets N_{1\subset2}+1$\;
        }
    }
    
    \tcp{Count points sampled in $s_2$ which are also inside $s_1$}
    \For{$x \in \calX_2$}{ 
        \uIf{$x \in s_1$}{
            $N_{2\subset1} \gets N_{2\subset1}+1$\;
        }
    }
    
    \tcp{Compute approximation of the intersection volume}
    Intersection $:= \dfrac{V_1 \cdot N_{2\subset1} + V_2 \cdot N_{1\subset2}}{2 N_{samples}}$\;
 \KwRet $\dfrac{\text{Intersection}}{\min(V_1, V_2)}$ \;}{} 

  \caption{The \FuncSty{isCompatible} function}
 \label{algo:box_intersect}
\end{algorithm}

\section{More About our Baselines}
\subsection{The Hill-Climbing Algorithm}
The Hill-Climbing algorithm~\cite{hillclimbing} is a naive greedy descent algorithm that constructs a solution iteratively, where at each iteration, it comprehensively searches for the proposal that best improves the objective function of a solution $\calS_F$, while leaving the solution valid \textit{i.e.} with no incompatibilities. If no proposal is available nor can improve the objective function, the algorithm stops \terminatesymb. A pseudo-code for the Hill-Climbing algorithm is provided in Algorithm~\ref{algo:UpHill}.


\begin{algorithm}[t]
\begin{multicols}{2}
\SetKwInOut{Input}{Input}
\SetKwProg{Init}{Initialize}{}{}
\SetAlgoLined
\SetKwFunction{iscompatible}{\text{isCompatible}}\SetKwFunction{evalloss}{\text{evalObjFunc}}
\SetKwProg{myproc}{procedure}{}{}
\KwResult{Set of selected \Cuboid $\calS_F$}
\Input{Set of proposal \Cuboid $\calS$\;}
 Threshold $\eta$ \;
 Final solution $\calS_F := \emptyset$\;
 Current best loss $\loss^* := +\infty$ \;
 Current best \Cuboid $s^* := \emptyset$\;
 Set of available \Cuboid $\calS_A := \calS$\;
 Evaluations counter $N_{eval} := 0$\;
\While{$\calS_A \neq \emptyset$}{
$s^*$ $\gets \emptyset$\;
\For{$s \in \calS_A$}{ 
    \uIf{ $s.$\iscompatible{$s_f$, $\eta$}}{
        $\calS_F$.add(s)\; 
        $\loss \gets$ \evalloss{$\calS_F$}\;
        $N_{eval} \gets N_{eval}+1$\;
        
        \uIf{$\loss < \loss^*$}{
            $\loss^* \gets \loss$\;
            $s^* \gets s$\;
        }
        $\calS_F$.remove($s$)\;
    }
    \Else{
    $\calS_A$.remove($s$)\;
  }
} 
$\calS_F$.add($s^*$)\; 
}
\KwRet $\calS_F$, $N_{eval}$
 \end{multicols}
 \caption{Hill-climbing algorithm}
 \label{algo:UpHill}
\end{algorithm}

\paragraph{Sub-optimal solutions} In practice, Hill-Climbing leads to sub-optimal solutions where the algorithm gets stuck into a local minimum. This is because the algorithm first greedily fits large regions of the scene, therefore employing large \Cuboid; This makes a lot of potentially good cuboids unavailable, as they would intersect with that large \Cuboid. 

\paragraph{Sub-optimality of evaluations}
Hill-climbing has to evaluate the complete objective function \evalsymb $\>\>$ \textit{each time} it considers a primitive, which is particularly costly at the beginning of the algorithm where the solution $\calS_F$ is still empty, since no \Cuboid candidate would intersect with it. After selecting a large \Cuboid, the set of available, \textit{i.e.} compatible cuboids gets dramatically reduced, hence resulting in an acceleration of the search of Hill-Climbing as shown in Figure~\ref{fig:progress_comparison}, which however converges to sub-optimal solutions. In contrast, MCTS and our algorithm evaluate the objective function only at the end of an iteration when a complete solution is complete.

\subsection{Binary-Tree MCTS}

\begin{figure}
    \centering
    \includegraphics[width=0.9\linewidth]{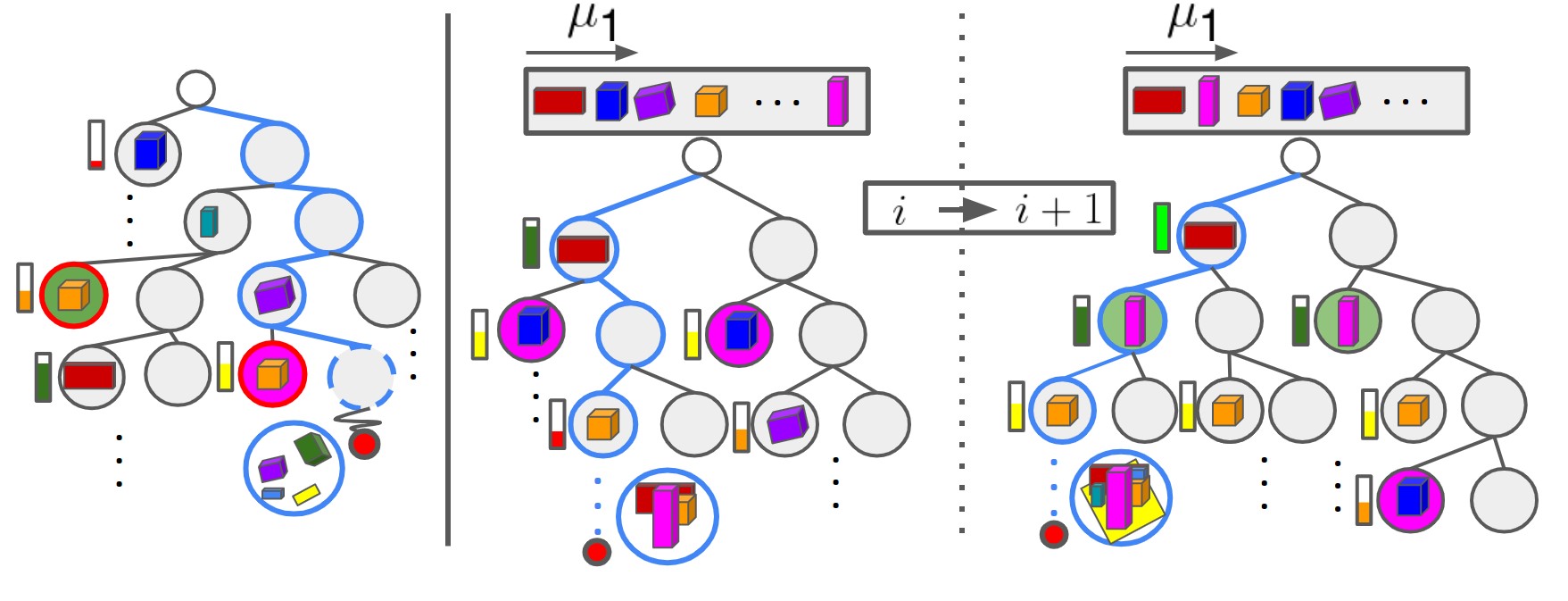}\\
    Binary-tree MCTS \hspace{0.6cm} 2 iterations of our algorithm interpreted with a binary tree
    \caption{We observe that behavior of our algorithm can be interpreted as an adaptive binary-tree MCTS, even though we do not explicitly define a tree structure. As MCTS is bound by its tree structure, it will invest iterations into exploring primitives in the upper part of the tree, even those with low confidence, visualized as \textbf{colored bars}. Further more, as indicated with \textbf{colored circles}, MCTS models confidences of same primitives in different parts of the tree independently. \textcolor{blue}{Blue circles} indicate a selection path of a single MCTS iteration that fails to extract meaningful proposals due to aforementioned difficulties. In contrast, our method sorts at each iteration primitives according to their confidences $\mu_1$ and will focus more easily on more promising primitives. In addition, as indicated by \textbf{colored circles} we only model a single confidence per primitive. These features enable our method to converge faster to good solutions in practice.}
    \label{fig:mcts2our}
\end{figure}

In our non-binary tree adaptation of MCTS, the search algorithm spends many iterations on iterating the first levels of the tree which might contain many, mutually incompatible, primitives. Therefore, this can limit the exploitation capabilities of MCTS as the algorithm prioritizes nodes that have not been visited yet. In our binary-tree adaptation of MCTS, every level of the tree corresponds to selecting or skipping a primitive. The resulting tree structure, hence, trades tree-breadth for tree-depth, which enables better exploitation. However, due to a large depth of the tree, MCTS does not explore solutions in the bottom of the tree, hence we observed only minor improvements over its non-binary adaptation. We argue that our MonteBoxFinder method can be seen as an adaptive version of binary-tree MCTS. In contrast to binary-tree MCTS, as we show in Figure~\ref{fig:mcts2our}, 
the tree equivalent of our method is able to adapt its structure during the search and enable better update mechanism leading to faster convergence.

\section{More results}
\subsection{Qualitative results}

We show more qualitative results in Figure~\ref{fig:qual_supp}.

\newlength{\niceresultsupwidth}
\setlength{\niceresultsupwidth}{0.28\linewidth}
\newcommand{\niceresultsup}[5]{
\includegraphics[width=\niceresultsupwidth, trim= #2 #3 #4 #5, clip]{images/qual/#1_uh.png} & 
\includegraphics[width=\niceresultsupwidth, trim= #2 #3 #4 #5, clip]{images/qual/#1_mcts.png} &
\includegraphics[width=\niceresultsupwidth, trim= #2 #3 #4 #5, clip]{images/qual/#1_ucb.png} \\}


\begin{figure}[t]
    \centering
    \begin{tabular}{ccc}
    \niceresultsup{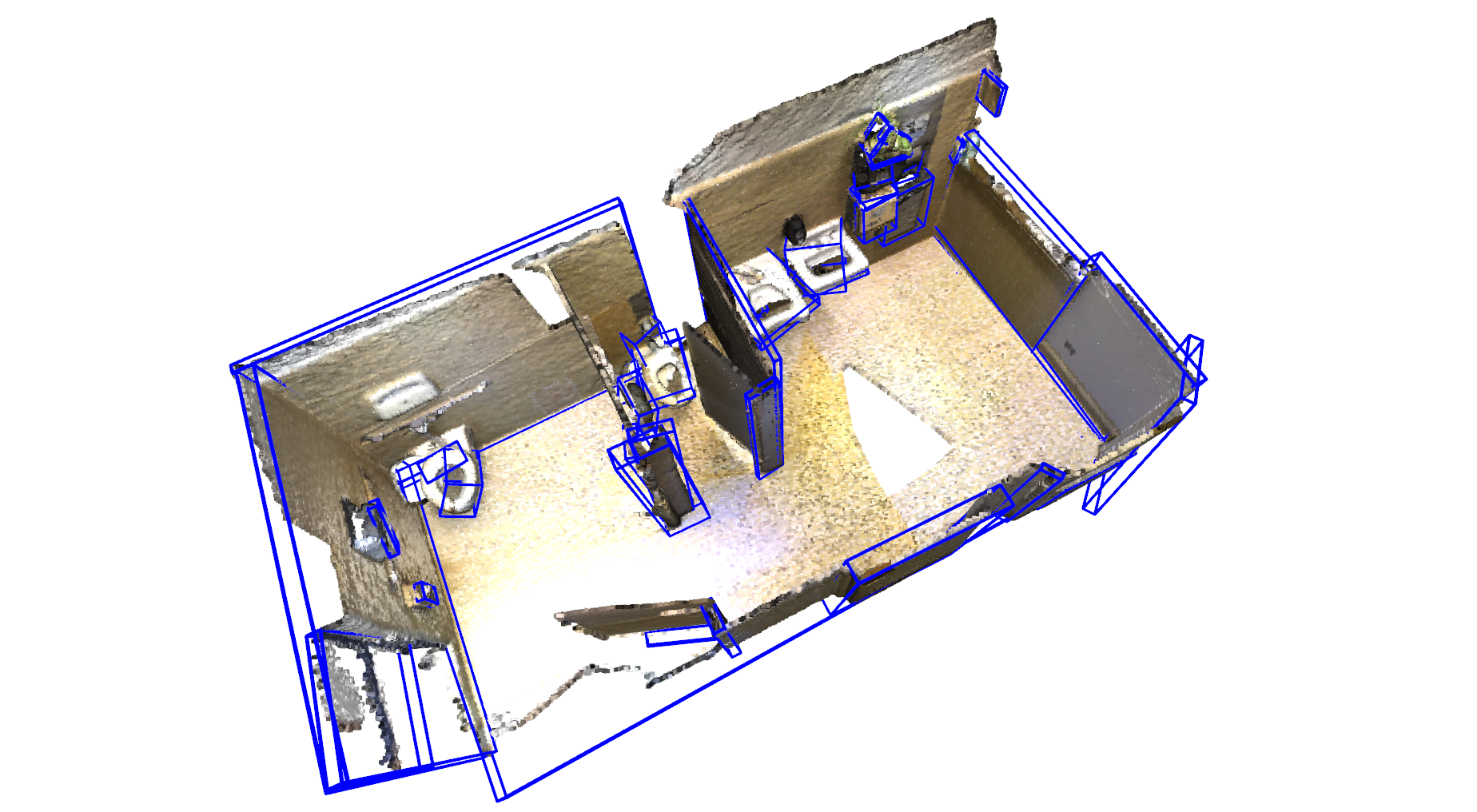}{300}{100}{500}{0}
    \niceresultsup{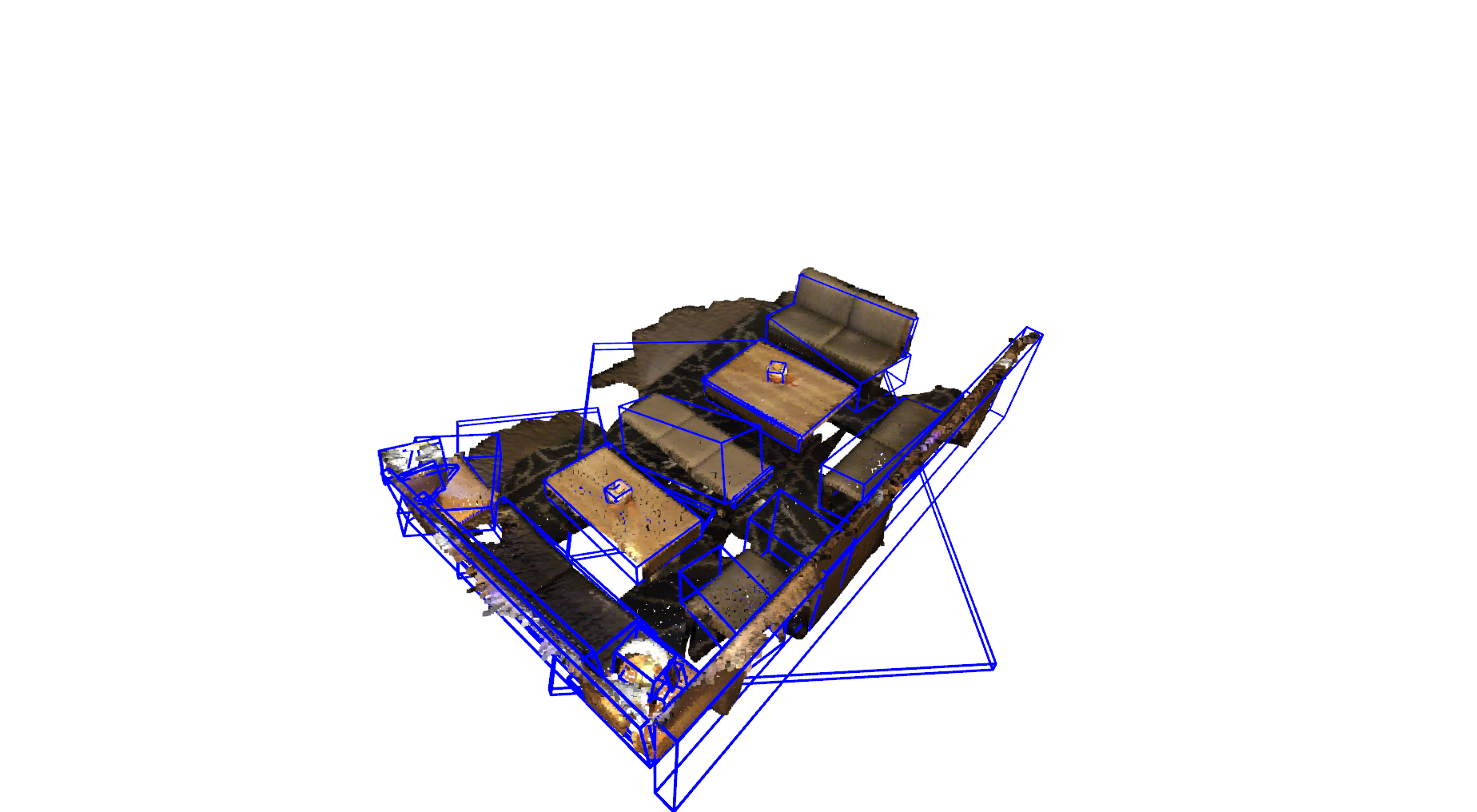}{500}{100}{500}{300}
    \niceresultsup{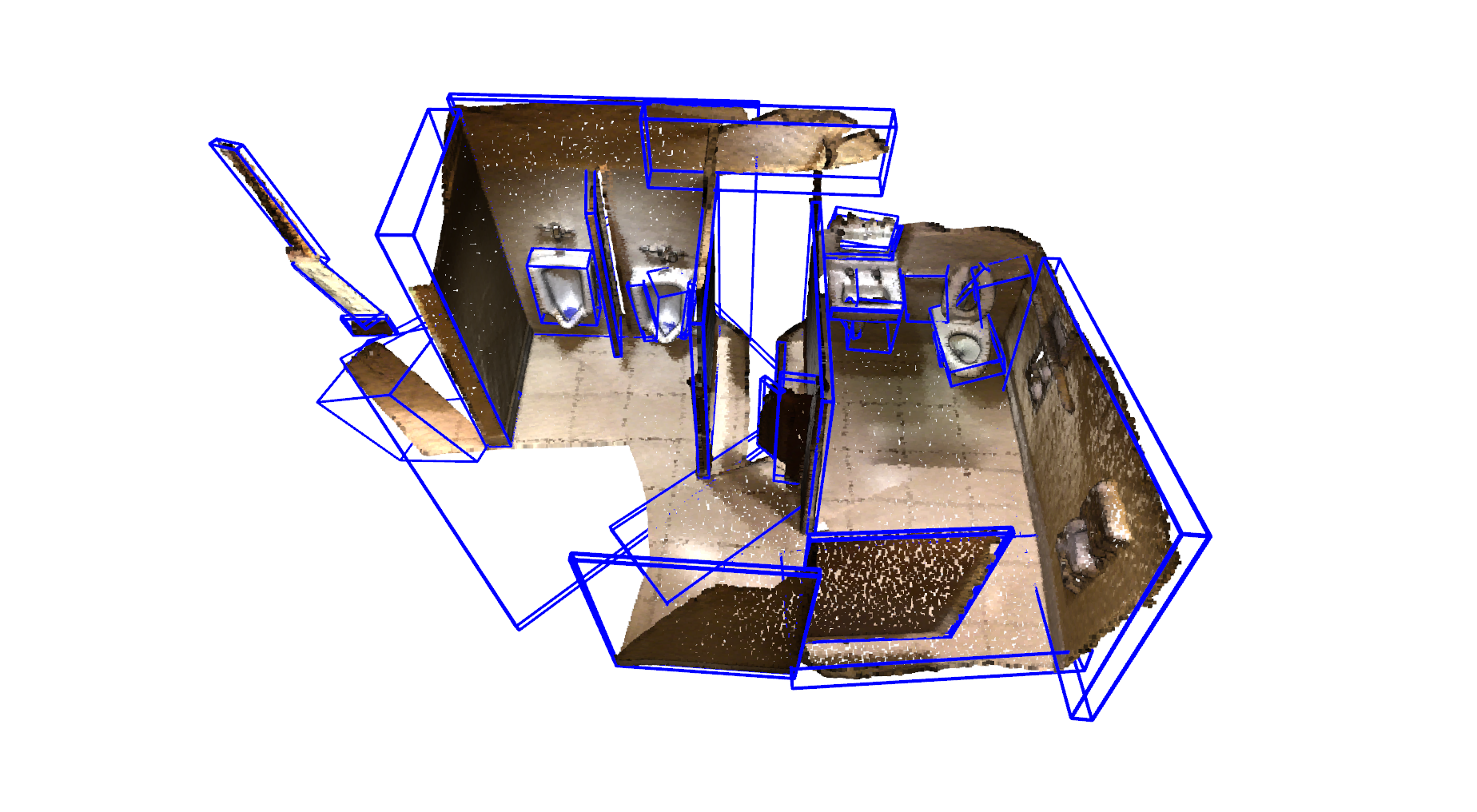}{300}{200}{300}{0}
    \niceresultsup{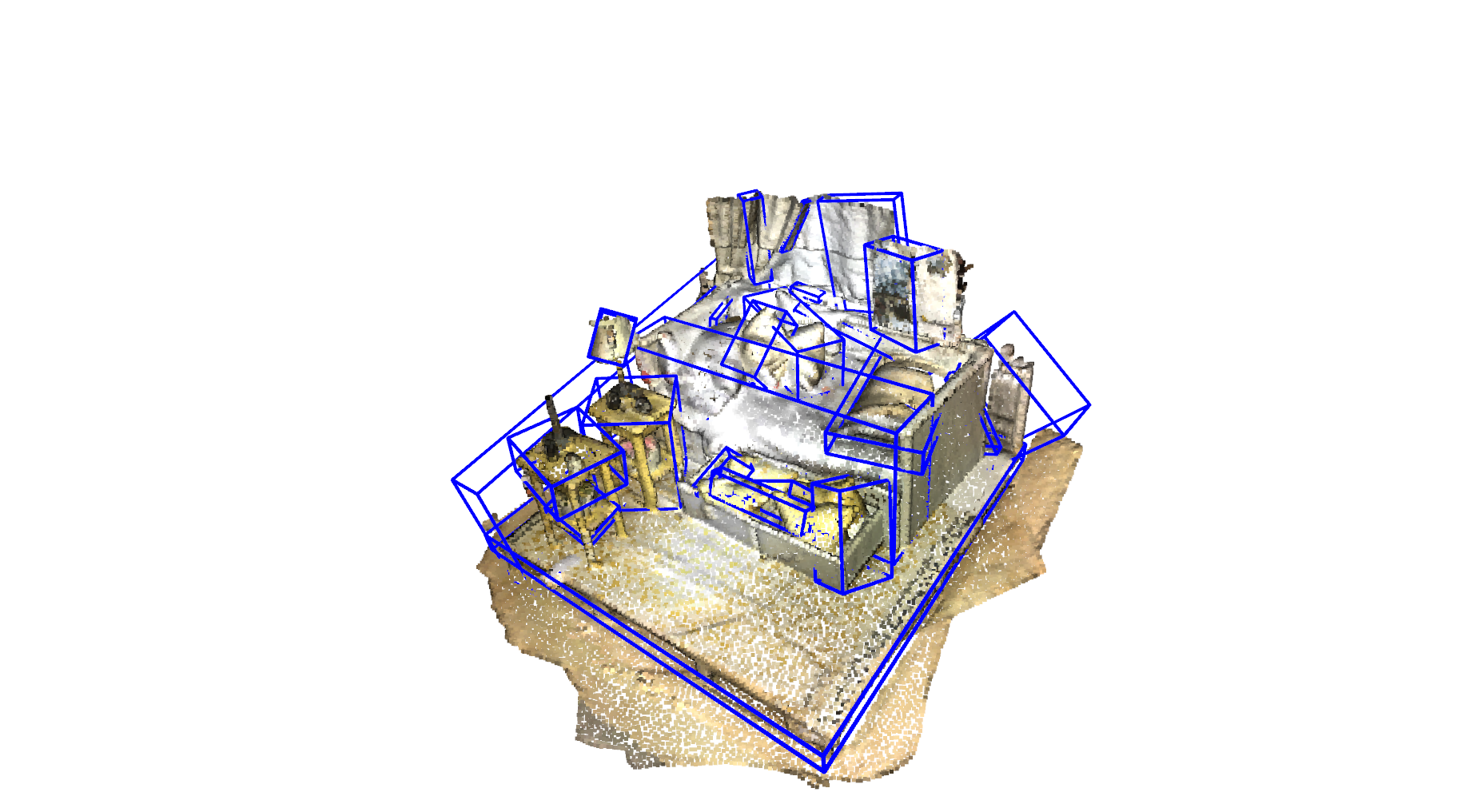}{500}{100}{500}{200}
    \niceresultsup{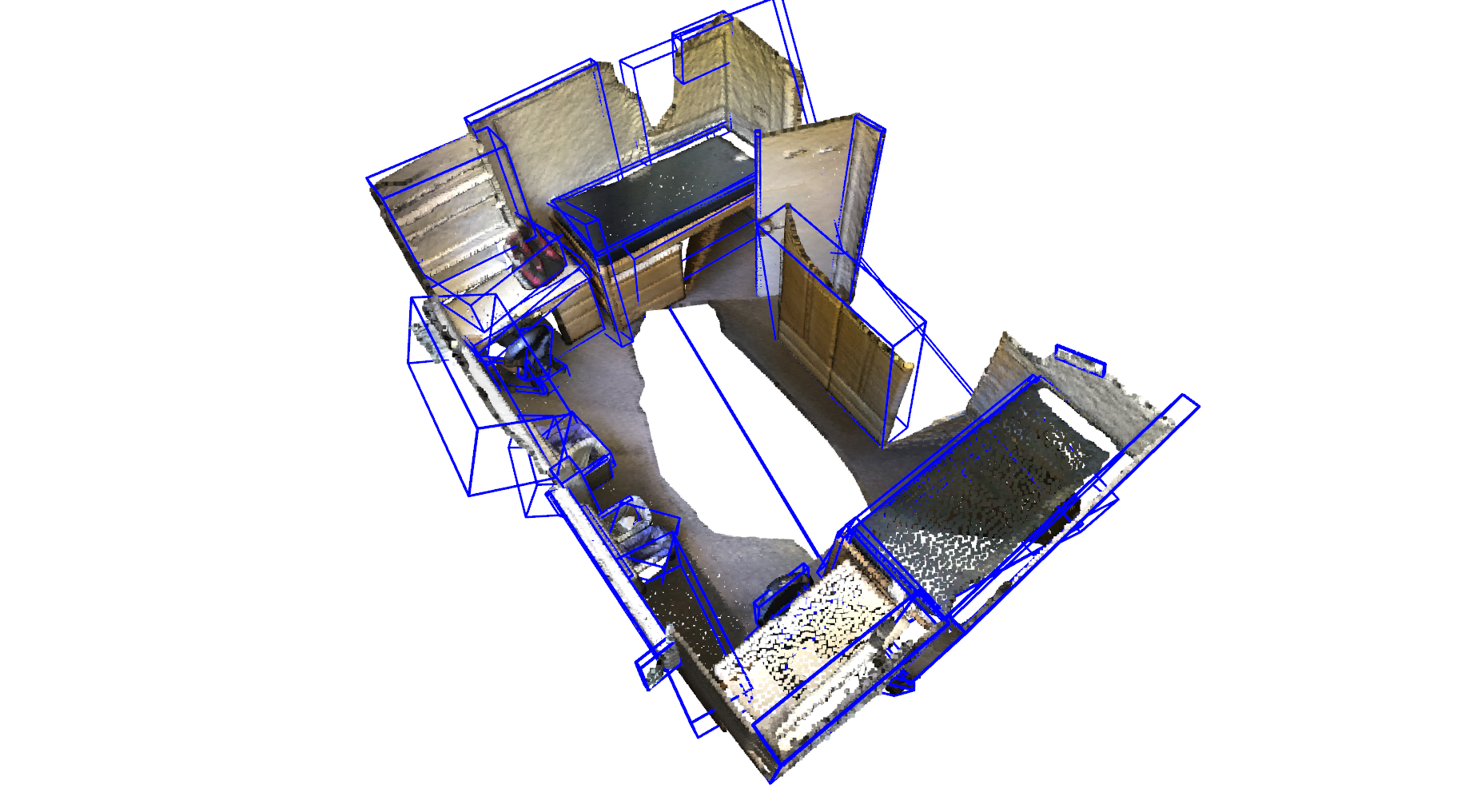}{500}{100}{400}{0}
    Hill-climbing & MCTS-Binary & MonteBoxFinder (ours) \\
    \end{tabular}
    \caption{\textbf{Qualitative results.} Hill-climbing often selects large cuboids that span across multiple different objects. MCTS does better, but sometimes yields outliers (second and fifth row). In contrast, our algorithm outperforms both methods and is able to successfully reconstruct many more details.}
    \label{fig:qual_supp}
\end{figure}

\subsection{Progress plots}

We show more progress plots in Figure~\ref{fig:progress_plot_supp}.

\newlength{\niceresultprogresswidth}
\setlength{\niceresultprogresswidth}{0.45\linewidth}
\newcommand{\niceresultprogress}[2]{
\includegraphics[width=\niceresultprogresswidth]{images/#1_progress.png} & 
\includegraphics[width=\niceresultprogresswidth]{images/#2_progress.png} \\} 


\begin{figure}[t]
    \centering
    \begin{tabular}{cc}
    \includegraphics[width=\niceresultprogresswidth]{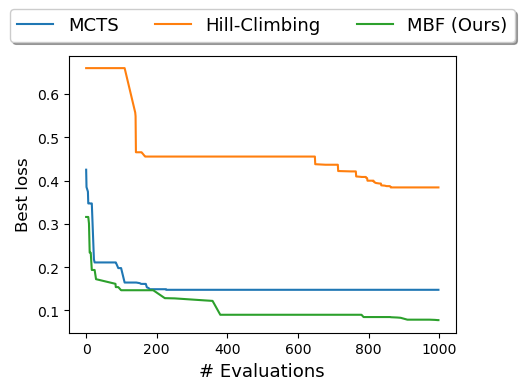} & 
    \includegraphics[width=\niceresultprogresswidth]{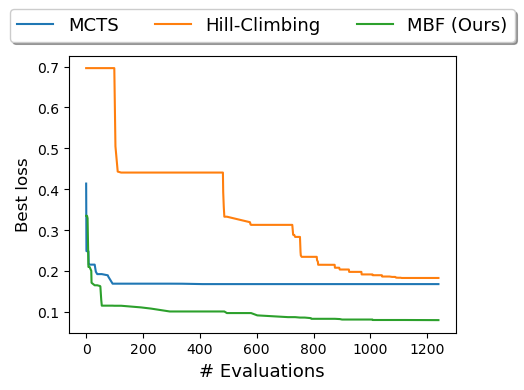} \\
    \includegraphics[width=\niceresultprogresswidth]{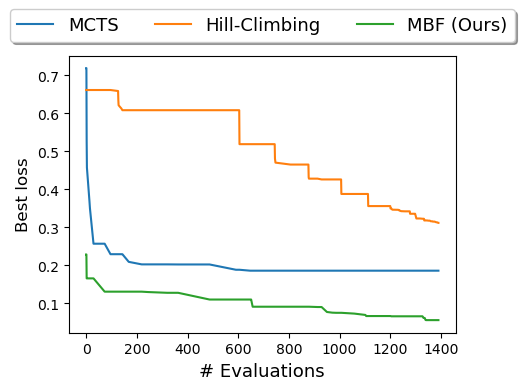} & 
    \includegraphics[width=\niceresultprogresswidth]{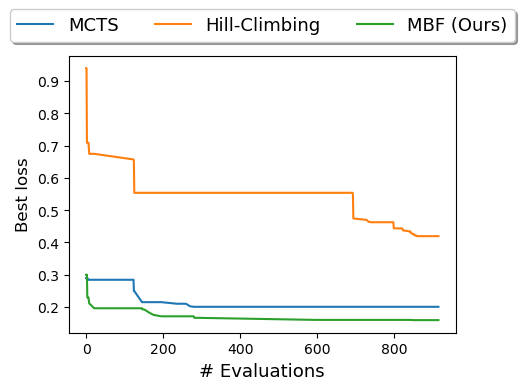} \\
    \includegraphics[width=\niceresultprogresswidth]{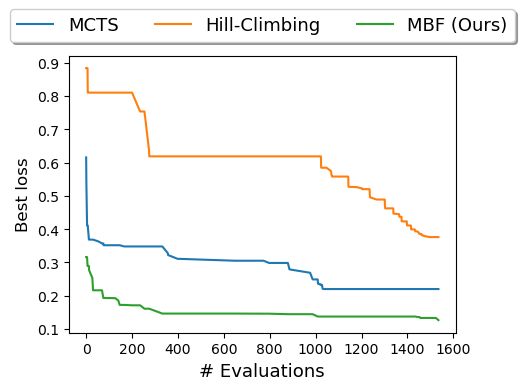} & 
    \\
    \end{tabular}
    \caption{\textbf{Samples of progress plots.} Our method consistently outperforms its baselines, i.e. the Hill-Climbing algorithm and MCTS, as it converges faster to a better solution. These plots correspond to the scene examples in Figure~\ref{fig:qual_supp}}
    \label{fig:progress_plot_supp}
\end{figure}

\end{document}